\documentclass[pdflatex,sn-mathphys-num]{sn-jnl}

\usepackage{url}
\usepackage{hyperref}
\usepackage{tcolorbox}

\usepackage{graphicx}%
\usepackage{multirow}%
\usepackage{amsmath,amssymb,amsfonts}%
\usepackage{amsthm}%
\usepackage{mathrsfs}%
\usepackage{xcolor}%
\usepackage{textcomp}%
\usepackage{manyfoot}%
\usepackage{booktabs}%
\usepackage{algorithm}%
\usepackage{algorithmicx}%
\usepackage{algpseudocode}%
\usepackage{listings}%
\usepackage{amsmath}
\usepackage{booktabs}

\usepackage{lmodern}
\usepackage{amsmath,amsfonts,amssymb}
\usepackage{graphicx}
\usepackage{hyperref}
\usepackage{booktabs}
\usepackage{siunitx}
\usepackage{microtype} 
\usepackage{geometry}
\usepackage{listings}    
\usepackage{etoolbox}
\usepackage{cleveref}
\usepackage{mdframed}  
\usepackage{xurl}
\usepackage{xspace}
\usepackage{siunitx}
\usepackage{caption}
\usepackage{pgfplots}  
\usepackage{csquotes}  

\theoremstyle{thmstyleone}%
\theoremstyle{thmstyletwo}%

\theoremstyle{thmstylethree}%

\begin{document}

\title[Non-Linear Scoring Model for Translation Quality Evaluation]{Non-Linear Scoring Model for Translation Quality Evaluation}


\author*[1]{\fnm{Serge} \sur{Gladkoff}}\email{serge.gladkoff@logrusglobal.com}

\author*[2,3]{\fnm{Lifeng} \sur{Han}}\email{l.han@lumc.nl}

\author[4]{\fnm{Katerina} \sur{Gasova}}\email{katerina.gasova@gmail.com}

\affil[1]{\orgdiv{Logrus Global LLC}, \orgname{}, \orgaddress{\city{Pennsylvania}, \country{United States}}}

\affil[2]{\orgdiv{LIACS}, \orgname{Leiden University}, \orgaddress{\street{Gorlaeus Gebouw - BE-vleugel, Einsteinweg 55}, \city{Leiden}, \postcode{2333 CC}, \state{Zuid-Holland}, \country{the Netherland}}}

\affil[3]{\orgdiv{BDS}, \orgname{Leiden University Medical Center (LUMC)}, \orgaddress{\street{Albinusdreef 2}, \city{Leiden}, \postcode{2333 ZG}, \state{Zuid-Holland}, \country{the Netherland}}}

\affil[4]{\orgdiv{MQM Council}, \orgname{Global Quality Solution Strategist, Argos Multilingual}, \orgaddress{\country{Czechia}}}


\abstract{Analytic Translation Quality Evaluation (TQE) based on Multidimensional Quality Metrics (MQM) typically applies a \emph{linear} error-to-penalty mapping calibrated on reference samples of \num{1000}--\num{2000} words. However, extrapolating this scale to other sample lengths introduces systematic bias, over-penalizing short texts and under-penalizing long ones, and often conflicting with expert judgments.
Building on our Multi-Range framework \cite{GladkoffEtAl2024}, we propose a calibrated non-linear scoring model that better reflects perceived translation quality across variable-length samples. Empirical evidence from three large-scale production evaluation settings indicates that acceptable error counts grow \emph{logarithmically}, rather than linearly, with sample size. This trend aligns with psychophysical and cognitive accounts (e.g., the Weber--Fechner law and Cognitive Load Theory), which suggest diminishing marginal perceptual impact of errors alongside increasing cumulative cognitive burden.
We introduce a two-parameter tolerance model anchored to a reference point and calibrated from two additional tolerance points via a one-dimensional root-finding procedure. The resulting formulation provides a closed-form range in which the linear approximation remains within $\pm 20\%$ relative error, and integrates into existing MQM workflows with minimal changes by replacing constant tolerance with a length-dependent function. The proposed method improves interpretability, fairness, and inter-rater reliability across both human and AI-generated translations, supporting more scalable and perceptually grounded translation quality evaluation.
}




\maketitle
\section{Introduction}
Analytic Translation Quality Evaluation (TQE) is widely regarded as a high-precision approach to assessing translation quality. It is based on systematic annotation of linguistic and technical errors using a predefined taxonomy, most commonly derived from the Multidimensional Quality Metrics (MQM) framework \cite{mqm2014,GladkoffEtAl2024}, and supported by structured annotator training. Once annotations have been produced, a scoring procedure is applied to convert error patterns into a quantitative quality estimate.

In operational settings, most scoring schemes adopt an implicit \emph{linear} scaling assumption: the number of permissible errors is expected to increase proportionally with the word count of the evaluated sample. In practice, this typically involves normalizing error counts to a reference sample size (e.g., 1{,}000 or 2{,}000 words) and comparing them against a user-defined threshold, such as allowing no more than five minor errors per 1{,}000 words. The resulting normalized error density is then used either to make an acceptability decision (pass/fail) or to derive a continuous quality score.\footnote{Logrus Global Statistics \url{https://logrusglobal.substack.com/archive}.}

While such linear normalization can be adequate for sample lengths near the calibration reference, it introduces systematic distortions when applied to substantially shorter or longer texts \cite{gladkoff2021}. Practitioners and researchers have repeatedly reported mismatches between linear penalty-based scoring and holistic judgments of translation quality. For instance, Waddington’s comparative studies \cite{waddington1999,waddington2001} show that additive points-off models alone correlate poorly with holistic evaluations, whereas combined analytic--holistic approaches can improve agreement. Similar effects have been documented in related evaluation domains. In speech and video quality assessment, international standards observe saturation effects in opinion scores and recommend comparison-based designs and \emph{non-linear mappings} (typically logistic) between objective indices and Mean Opinion Scores (MOS) \cite{ituP800,streijl2016,vqeglogistic}. Within translation quality assurance practice, non-linearity is also reflected through severity weighting and the treatment of critical ``showstopper'' errors, where a single major defect may outweigh many minor ones \cite{mqm2014,saeJ2450}. Large-scale machine translation evaluations further show that expert, context-aware severity judgments yield rankings that diverge from simple averaging, providing additional evidence for non-additive scoring \cite{freitag2021}. Related patterns are well established in usability engineering, where a single catastrophic defect can dominate acceptability decisions \cite{nielsen1994}. Collectively, these findings suggest that perceived translation quality is not well characterized by linear accumulation of penalties.

Despite these known limitations, linear error scaling remains common in industry practice, partly due to institutional convention and partly due to the tendency to apply linear extrapolation in numerical reasoning. Importantly, when evaluators assess translations without explicitly performing normalization, their judgments often deviate from what a linear model would predict.\footnote{We provide examples of institutional user reports in Appendix~\ref{Appendix_users_quote}.}



This motivates the research questions addressed in this work:

\begin{quote}
\textbf{RQ1:} \emph{How does the maximum number of tolerable MQM errors change as a function of evaluation sample length in real-world translation quality assessment settings?}\\
\textbf{RQ2:} \emph{How can this relationship be operationalized as a calibrated non-linear scoring model that remains interpretable, fair across sample lengths, and compatible with existing MQM-based evaluation workflows?}
\end{quote}

To answer these questions, we extend our earlier Multi-Range TQE framework \cite{GladkoffEtAl2024} and argue that tolerance for errors increases with sample length, but at a diminishing rate; equivalently, the acceptable error rate per 1{,}000 words decreases as text length increases. We model this behavior using a logarithmic tolerance function and derive a practical calibration procedure that can be integrated into existing MQM workflows with minimal modification.

\paragraph{Contributions.}
This paper makes the following contributions:
\begin{itemize}
    \item We provide empirical evidence from three large-scale production evaluation settings addressing \textbf{RQ1}, showing that acceptable error counts grow \emph{sublinearly} with sample size, consistent with a logarithmic dependence.
    \item We link the observed scaling behavior to established psychophysical and cognitive theories, including the Weber--Fechner law and Cognitive Load Theory, thereby motivating a principled non-linear tolerance function.
    \item We propose a calibrated two-parameter non-linear tolerance model (\textbf{RQ2}) that can be fit from two tolerance points via a one-dimensional root-finding procedure, yielding a practical scoring function for variable-length evaluation.
    \item We show that the proposed model can be integrated into existing MQM-based analytic TQE workflows with minimal modification by replacing constant error tolerance with a length-dependent tolerance function, improving interpretability and supporting more consistent evaluation across sample sizes.
\end{itemize}

The remainder of this paper is structured as follows. Section~\ref{sec:background} reviews analytic TQE and MQM-based scoring. Section~\ref{sec:empirical} presents empirical evidence for non-linear tolerance growth. Section~\ref{sec:physhophysical} discusses psychophysical and cognitive foundations. Section~\ref{sec:NLS} introduces the proposed non-linear scoring model. Sections~\ref{sec:practical} and~\ref{sec:building} discuss practical implications and provide guidance for constructing a non-linear scorecard. Section~\ref{sec:limitation} concludes with limitations and directions for future work.

\section{Background}
\label{sec:background}

\subsection{Analytic TQE and MQM}

Analytic Translation Quality Evaluation (TQE) is a structured approach to assessing translation output by identifying, categorizing, and recording observable errors with respect to the source text and relevant specifications. In contrast to holistic evaluation methods, which produce a single overall impression of quality, analytic TQE requires systematic annotation of discrete error instances, each assigned to a predefined error type and (typically) a severity level.

The most widely adopted framework for such structured error annotation is the Multidimensional Quality Metrics (MQM) standard \cite{mqm2014}. MQM defines a hierarchical, extensible typology with high-level error dimensions such as \emph{Accuracy} and \emph{Linguistic Conventions} (formerly \emph{Fluency}), and more fine-grained categories including \emph{Mistranslation}, \emph{Omission}, \emph{Grammar}, and \emph{Spelling}. In practice, many industry evaluation schemes implement MQM directly or adopt customized subsets tailored to specific content types, client requirements, or quality objectives.

Following annotation, a scoring procedure is applied to an evaluation sample---a defined portion of a translation job---to convert the set of annotated errors into either (i)~a numeric quality score or (ii)~an acceptability decision (pass/fail). MQM-based evaluation therefore consists of two complementary components: \emph{(i)~an error typology} specifying which error categories and severity levels are annotated, and \emph{(ii)~a scoring model} that maps these annotations to severity-weighted penalty points and an overall score. For comparability across texts of different lengths, scores are typically expressed relative to a fixed reference sample size, most commonly 1{,}000--2{,}000 words.\footnote{\url{https://themqm.org/}.}

\subsection{Linear Scoring Models and Their Limits}

The dominant scoring approach in analytic TQE is based on \emph{linear length normalization}. Under this assumption, the \emph{severity-weighted penalty total} (\textsc{APT}, Absolute Penalty Total; i.e., the sum of penalty points assigned to annotated errors) is scaled proportionally to a reference word count so that results can be compared across evaluation samples of different sizes. For example, a tolerance of five minor errors (one point each) per 1{,}000 words corresponds to 2.5 penalty points in a 500-word sample and 10 penalty points in a 2{,}000-word sample. If major errors receive a $5\times$ severity multiplier, then a single major error consumes the full five-point allowance for a 1{,}000-word sample.

Although proportional scaling provides a simple and operationally convenient normalization rule, it does not fully capture how translation quality is perceived in practice. Empirical and practitioner-oriented work suggests that evaluators apply stricter expectations to very short texts, where individual errors are salient, while showing greater tolerance for longer passages \cite{waddington1999,waddington2001,mqm2014,saeJ2450}. At the same time, tolerance does not increase indefinitely with length: as documents grow, error accumulation and compounding effects contribute to declining acceptability. For instance, a single major error may be sufficient to fail a one-page sample, yet a seven-page sample would not typically be judged acceptable with seven major errors, even though proportional scaling would imply this threshold. Such judgments indicate that the relationship between sample length and acceptable error count is fundamentally non-linear.

Building on the Multi-Range framework proposed in \cite{GladkoffEtAl2024}, we distinguish three regimes of evaluation sample size:

\begin{itemize}
  \item \textbf{Micro-range} (\(<250\) words): For very short samples, analytic scoring exhibits high sampling uncertainty. As shown in \cite{gladkoff2021}, estimates derived from such brief segments yield wide confidence intervals, limiting their interpretability for operational decision-making.

  This limitation is also relevant for sentence- or segment-level Quality Estimation (QE). Recent industry observations suggest that fine-grained QE predictions may not consistently reflect document-level acceptability, particularly when critical errors occur sparsely or depend on broader context \cite{baneATA2025,gladkoff2025substack}. These findings motivate treating micro-range scoring as a distinct regime in which statistical uncertainty dominates, and where Statistical Quality Control (SQC) methods can provide a more risk-aware interpretation of limited observations.



  \item \textbf{Meso-range} (\(\approx250\!-\!3{,}000\) words): Within this range, scoring based on a reference-length normalization can be operationally effective. In particular, a \emph{linearly calibrated scoring model} anchored to a user-defined tolerance in severity-weighted penalty points (APT) at a reference word count is commonly used.

  \item \textbf{Macro-range} (\(>3{,}000\) words): For longer samples, human judgments increasingly diverge from linear extrapolation calibrated in the meso-range. In this regime, a \emph{non-linear} tolerance function---in particular, a logarithmic scaling curve---is better suited to model perceived quality across both meso- and macro-range texts without requiring multiple piecewise linear approximations.
\end{itemize}

The present study focuses on the meso- and macro-range, where analytic measurements are statistically more stable and practically meaningful. Here, ``statistically reliable'' refers to estimates with sufficiently narrow confidence intervals, i.e., sampling-uncertainty bounds that shrink as the evaluated sample size increases.\footnote{For intuition, a simple 95\% Wald interval for a proportion is \(p \pm 1.96\sqrt{p(1-p)/n}\) \cite{gladkoff2021}. For small samples or extreme proportions, this estimate is known to undercover; a \emph{Wilson (score)} interval---or the closely related \emph{Agresti--Coull} ``plus-four'' approximation---provides improved coverage \cite{AgrestiCoull1998}. \emph{Wilson 95\% CI: }
\(\displaystyle
\frac{\hat p+\frac{z^2}{2n} \pm z\sqrt{\frac{\hat p(1-\hat p)}{n}+\frac{z^2}{4n^2}}}{1+\frac{z^2}{n}},
\)
with \(z\!=\!1.96\); the Agresti--Coull ``plus-four'' uses \(\tilde p=(x+\tfrac{z^2}{2})/(n+z^2)\) and
\(\tilde p \pm z\sqrt{\tfrac{\tilde p(1-\tilde p)}{n+z^2}}\) \cite{AgrestiCoull1998}.
}

Applying linear scoring indiscriminately across all sample sizes often leads to systematic friction in evaluation practice. Users and linguists may need to override, reinterpret, or replace scores, or maintain multiple length-specific scoring rules for short and long samples. A common workaround is to restrict evaluation to a fixed 1{,}000-word segment; however, this strategy is operationally restrictive and does not address the underlying mismatch between linear scaling and perceived quality. Moreover, because confidence interval width decreases approximately as $1/\sqrt{n}$---with an additional finite-population correction when the evaluated sample constitutes a substantial fraction of the document---variation in sampled word count can materially affect estimation uncertainty.\footnote{When the sample is a large fraction of the document, apply the finite-population factor $\sqrt{(N-n)/(N-1)}$ \cite{gladkoff2021} compared to the simple 95\% Wald form of $p \pm 1.96\sqrt{p(1-p)/n}$.}

The persistence of linear scoring practices, despite these limitations, can be attributed in part to the lack of simple, theoretically grounded alternatives that can be deployed within established MQM workflows. The present work addresses this gap by proposing a calibrated non-linear scoring model, supported by both empirical evidence and theoretical motivation drawn from psychophysics and cognitive-load research.

\section{Empirical Evidence of Non-Linearity}
\label{sec:empirical}

To examine how perceived translation quality varies with evaluation sample length, we conducted an empirical elicitation study with Quality Managers from three large institutional translation programs (anonymized). Each organization operates a high-volume multilingual workflow and applies mature, metric-driven TQE procedures to support routine quality monitoring and acceptance decisions.

\subsection{Survey Design}

We reused the \emph{extended calibration questionnaire} introduced in \cite{GladkoffEtAl2024}. The instrument elicits respondents’ tolerance judgments for multiple sample sizes while explicitly discouraging linear extrapolation, thereby enabling the recovery of latent non-linear tolerance functions.

The questionnaire was designed to capture Quality Managers’ intuitive notion of \emph{error tolerance}, operationalized as the maximum number of severity-weighted penalty points (\textsc{APT}) that would still be considered acceptable for a given sample size. The key methodological challenge was controlling for respondents’ prior exposure to standard length-normalized scoring schemes, which can lead professionals to default to proportional mental scaling. To mitigate such anchoring effects, respondents were instructed not to compute tolerances from existing formulas or reference thresholds, but instead to report what their organization would consider acceptable based on professional judgment and operational experience.

The questionnaire contained the following prompts:

\begin{tcolorbox}[title={Intuitive Error Tolerance Questions}]
\begin{itemize}
    \item Intuitively, how many minor errors would you accept in a 1-page sample?
    \item Without doing any math or using a scoring model, what number of minor errors feels acceptable for 2 pages?
    \item For a 3-page text, how many minor errors seem tolerable---based on your experience?
    \item Do not calculate---just estimate: how many minor errors would you allow in 4 pages?
    \item Based on your intuition, how many minor errors would still be acceptable in a 5-page sample?
    \item Without applying your formal model, what number of minor errors seems acceptable in 10 pages?
    \item Without referencing formulas, what minor error count would feel acceptable for 20 pages?
\end{itemize}
\end{tcolorbox}

\paragraph{Elicitation controls.}
To reduce anchoring and promote independent judgments across lengths, page sizes were presented in randomized order and prior responses were not displayed. Two page sizes were repeated at the end of the questionnaire to probe intra-rater consistency. Respondents were again explicitly instructed not to prorate from known thresholds and to rely on professional judgment rather than calculation.

The elicitation focused on \emph{minor} errors for two reasons. First, higher-severity errors (major and critical) occur relatively infrequently, which makes them less suitable for robust quantitative elicitation at the sample sizes considered. Second, their perceptual and operational impact differs qualitatively, as a small number of high-severity defects may dominate acceptability decisions. In industry practice, severity-weighted scoring allows the effect of any error to be expressed as an equivalent number of minor errors, making minor-error tolerance a practical proxy for overall tolerance modeling.

Respondents expressed sample length either in pages or in word counts. For comparability across institutions, all answers were normalized to pages using a conversion factor of 250 words per page.

\subsection{Results}

Across all three institutions, responses exhibited a consistent non-linear tolerance pattern. Respondents indicated that longer samples permit a higher absolute number of errors, but that this permissible count grows substantially more slowly than sample length. Empirically, tolerance increased steeply for short samples and gradually saturated for larger samples. This qualitative shape was observed despite differences in each institution’s baseline standards, performance targets, and content types.

\begin{figure*}[t]
  \centering
  \includegraphics[width=0.69\linewidth]{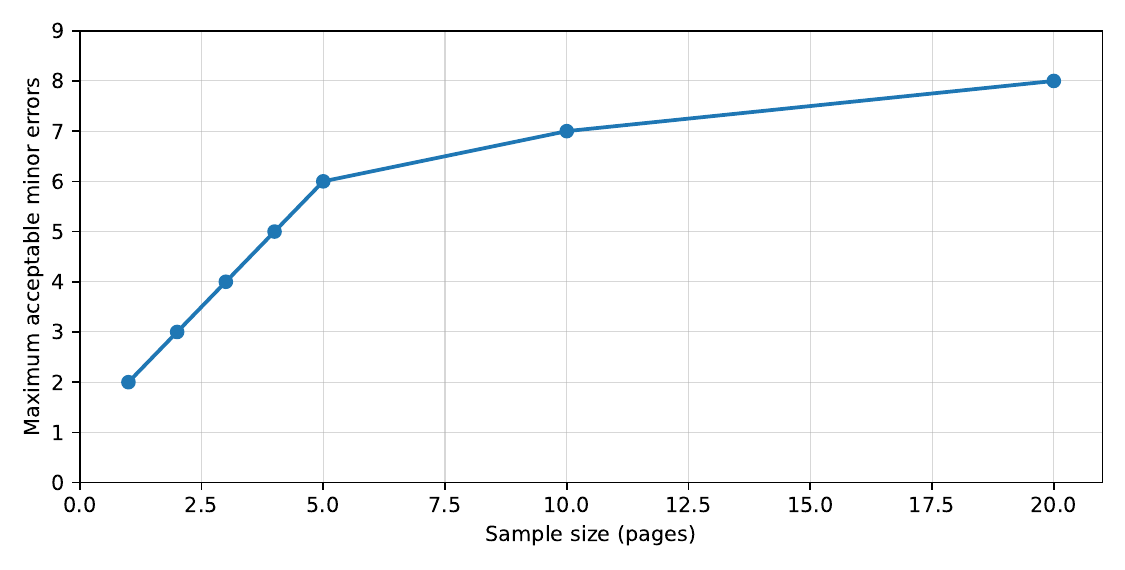}
  \caption{\textbf{Institution~1 raw questionnaire responses.} X-axis: sample size in pages (\(250\)~words/page); Y-axis: maximum acceptable number of \emph{minor} errors.}
  \label{fig:client1}
\end{figure*}

\begin{figure*}[t]
  \centering
  \includegraphics[width=0.99\linewidth]{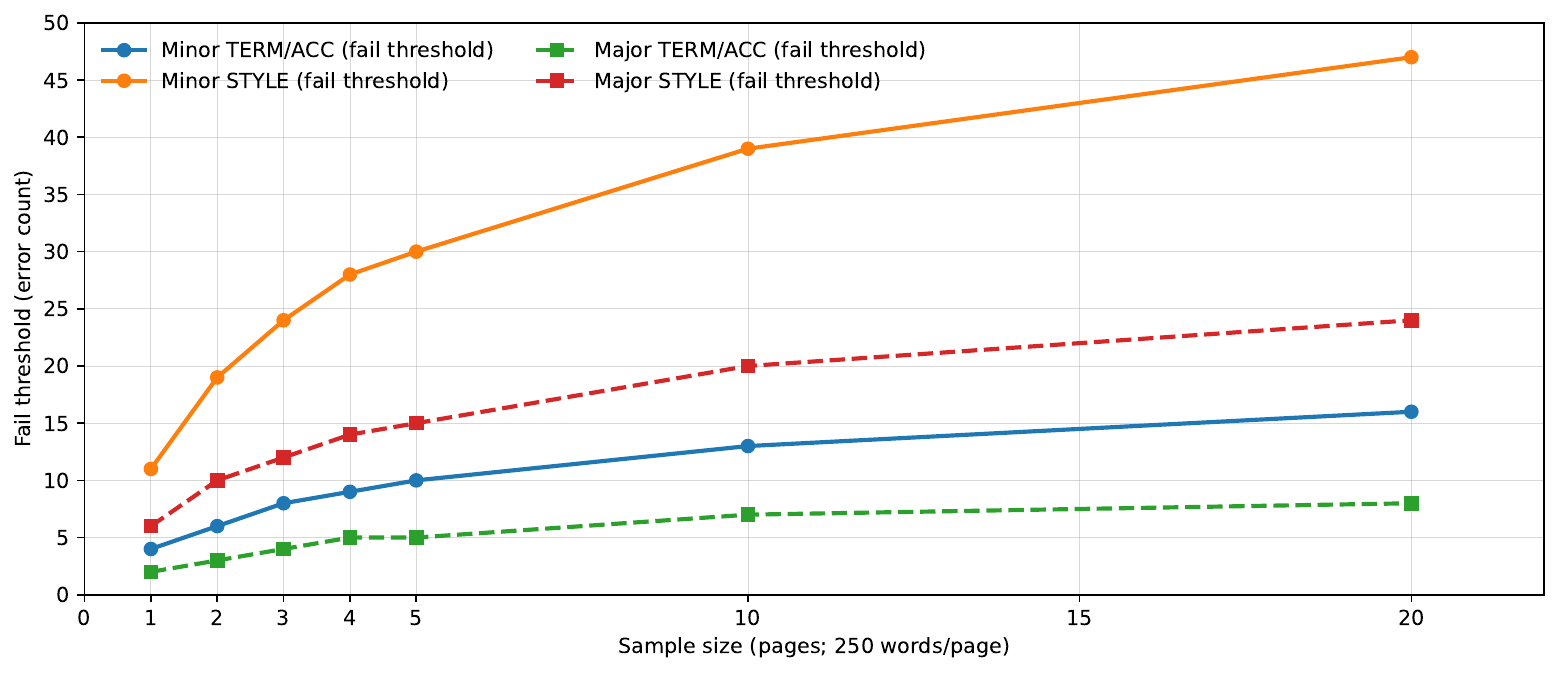}
  \caption{\textbf{Institution~2 raw questionnaire responses.} X-axis: sample size in pages (\(250\)~words/page). Y-axis: fail thresholds (maximum acceptable counts) for four elicited series: minor and major errors, each for TERM/ACC and STYLE categories.}
  \label{fig:client2}
\end{figure*}

\begin{figure*}[t]
  \includegraphics[width=0.99\linewidth]{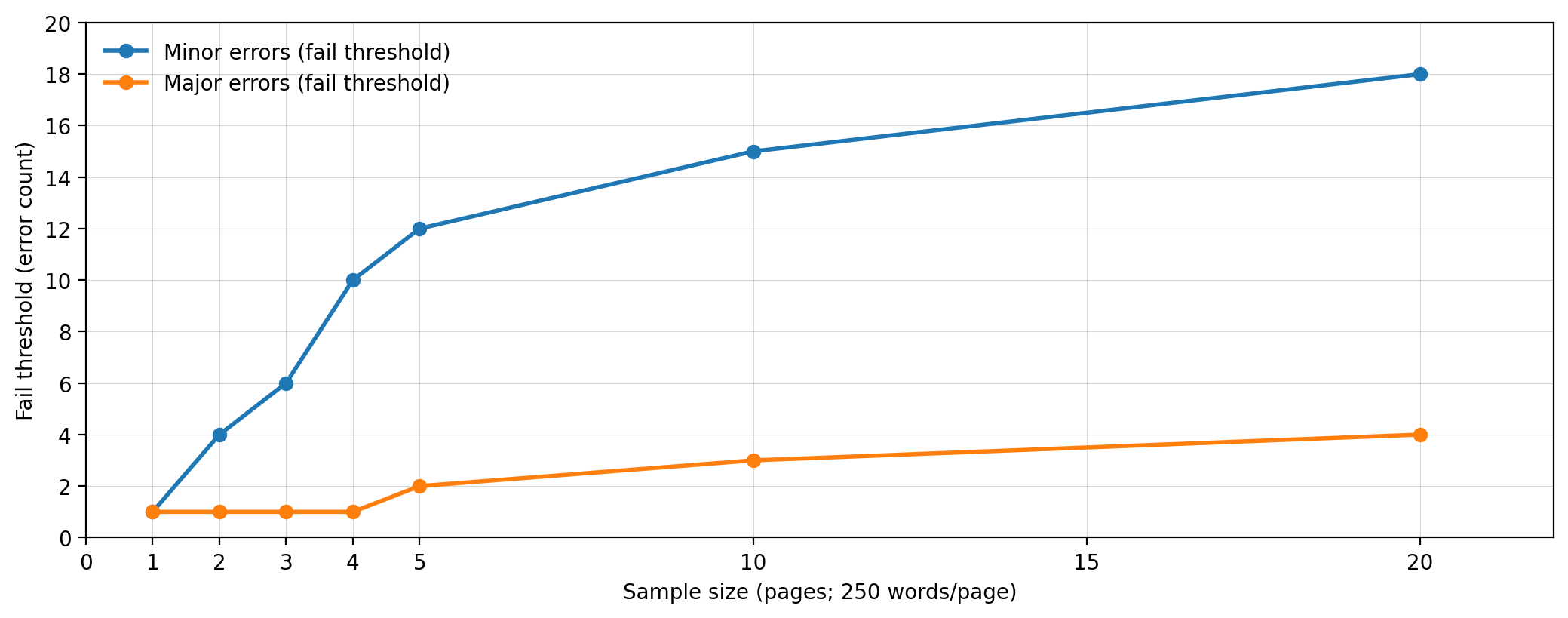}
  \caption{\textbf{Institution~3 raw questionnaire responses.} X-axis: pages (\(250\)~words/page); Y-axis: maximum allowed number of errors (two series shown, \emph{minor} and \emph{major} errors).}
  \label{fig:client3}
\end{figure*}

Figures~\ref{fig:client1}--\ref{fig:client3} report the raw elicitation results. The horizontal axis represents sample length in pages and the vertical axis represents the maximum number of errors considered acceptable. In one institution, tolerance judgments were collected for both minor and major error series.

None of the observed tolerance functions were well described by a straight-line relationship. Although institutions differed in absolute tolerance level, the response curves shared a similar concave shape and were well approximated by a logarithmic form. Notably, respondents reported that the elicitation process itself made the non-linearity salient: when asked to suspend explicit calculation, they recognized that their implicit acceptance criteria assign disproportionately high weight to errors in short samples, while additional errors in longer texts contribute less marginal impact.
These results suggest that acceptance judgments depend not only on the absolute number of errors, but also on how errors accumulate across increasing text length. As a consequence, tolerance does not scale proportionally with sample size, resulting in non-linear growth in acceptable error counts.

Consistent with the pilot results reported in \cite{GladkoffEtAl2024}, the present findings indicate that linear scaling substantially overestimates tolerable error counts in the macro-range, in some cases by up to 50\%, whereas a logarithmic tolerance function tracks expert judgments more closely over the 1--20 page range. Based on this convergent pattern, we hypothesize that human tolerance for translation errors grows approximately logarithmically with sample size. The remainder of this paper examines theoretical explanations for this relationship and introduces a calibrated scoring model that operationalizes it within MQM-based evaluation practice.

From Figures~\ref{fig:client1}--\ref{fig:client3}, it is evident that the elicited tolerance points do not follow a linear trend. For readers interested in formal model comparison, Appendix~\ref{appendix:linear-vs-logarithmic} provides quantitative details on linear versus logarithmic fits, including the results underlying the trend observed in Figure~\ref{fig:client3}.

We note that these results reflect expert elicitation rather than controlled behavioral experiments; nevertheless, the consistent concave trend across institutions provides strong evidence of non-linear tolerance scaling in operational settings.

\section{Psychophysical and Cognitive Foundations}
\label{sec:physhophysical}

The elicitation results in Section~\ref{sec:empirical} suggest that perceived translation quality does not vary linearly with either sample length or the total number of annotated errors. This section provides theoretical motivation for such non-linearity by drawing on two complementary perspectives: classical psychophysical scaling and Cognitive Load Theory. We additionally relate these accounts to document-level MT evaluation findings, where context and aggregation are known to influence human judgments and agreement patterns \cite{castilho-2020-page,grundkiewicz-etal-2021-user,vernikos-etal-2022-embarrassingly,kim-2025-context}.

\paragraph{Scope and interpretive caveat.}
Translation errors are not physical stimuli in the strict psychophysical sense, and we do not claim that quality judgments obey a literal Weber--Fechner law. Rather, psychophysics and cognitive theories offer principled \emph{functional} explanations for why human sensitivity to additional defects is typically compressive and why aggregation over longer texts can yield length-dependent acceptability thresholds. These perspectives provide a theoretically grounded motivation for replacing proportional (linear) extrapolation with a calibrated sublinear tolerance function in MQM-style analytic scoring.

\subsection{Weber--Fechner Law and Logarithmic Perception}

Psychophysics has long established that subjective sensation typically grows non-linearly with stimulus intensity. Weber observed that the \emph{just-noticeable difference} (JND) is approximately proportional to the baseline stimulus magnitude \cite{weber1834}. Fechner subsequently formalized a quantitative model in which equal stimulus ratios correspond to equal increments in sensation, implying a logarithmic relationship between physical intensity and perceived magnitude \cite{fechner1860}. This principle, now commonly referred to as the Weber--Fechner law, is a canonical example of compressive perceptual scaling (see also \cite{boring1950,fechner1966}).

\paragraph{Implication for TQE.}
As an analogy for translation quality evaluation, compressive scaling predicts diminishing marginal perceptual impact of additional errors. That is, the subjective penalty associated with one more defect is expected to be larger when the text is nearly error-free and smaller when multiple errors have already been encountered. This prediction is consistent with the empirical pattern observed in Section~\ref{sec:empirical}, where tolerance rises rapidly for short samples and flattens as sample length increases.

Subsequent psychophysical work refined this relationship while preserving its central insight. Stevens proposed that many perceptual modalities follow a power function \(S = kI^{n}\), often with exponents \(n<1\), yielding concave response curves consistent with sublinear growth \cite{stevens1957}. Moreover, the logarithmic function can be viewed as a limiting case of a power law:
\[
\lim_{n\to 0}\frac{I^{\,n}-1}{n} \;=\; \ln I.
\]
Over practical dynamic ranges, small-exponent power functions can approximate logarithmic behavior closely. A logarithmic tolerance curve therefore provides a parsimonious functional form for modeling diminishing marginal sensitivity in quality judgments. Comprehensive psychophysical treatments and signal detection theory syntheses likewise emphasize compressive mappings as common across perceptual domains \cite{gescheider1997,green1966}.

Taken together, the psychophysical literature motivates the hypothesis that translation errors, treated as discrete ``quality-relevant events,'' are integrated by human evaluators using a compressive internal scale. This provides one theoretical justification for sublinear growth in acceptable error counts as evaluation samples become longer.

\subsection{Cognitive Load and Cumulative Disruption}

Whereas psychophysical scaling accounts for diminishing marginal sensitivity to repeated stimuli, cognitive theories emphasize how repeated disruptions impose increasing demands on limited processing resources. Cognitive Load Theory (CLT) formalizes the idea that working memory capacity is constrained and that comprehension degrades when cognitive demands exceed these limits \cite{sweller1988}, consistent with foundational observations on short-term memory capacity \cite{miller1956}.

In translation evaluation, each error can function as a localized disruption that requires attention, reinterpretation, or correction by the reader. While isolated minor errors may be tolerated, repeated disruptions can accumulate and reduce processing fluency, increasing perceived effort and lowering confidence in the text. As a result, acceptability can decline non-linearly with total error accumulation even when errors are individually low-severity. Related effects are documented in document-level MT evaluation: presenting broader context can change judgments and alter agreement patterns relative to sentence-level setups \cite{castilho-2020-page,grundkiewicz-etal-2021-user}, and document-level aggregation can capture quality factors that are not reliably reflected by sentence-level scoring alone \cite{vernikos-etal-2022-embarrassingly}. Recent analyses further suggest that ``context'' effects may be pervasive yet heterogeneous, complicating attempts to infer document-level acceptability from local judgments \cite{kim-2025-context}.

CLT also provides a plausible interpretation of length effects reported in applied translation assessment: as the expected cognitive demands of processing a long document increase, evaluators may adopt stricter acceptance thresholds to reduce the risk of cumulative disruption, consistent with divergences between analytic scoring and holistic judgments in longer samples \cite{waddington1999,waddington2001}.

\subsection{Implications for Length-Dependent Tolerance Modeling}

In summary, psychophysical scaling and cognitive-load considerations converge on a shared expectation: tolerance should increase with sample length, but at a diminishing rate. A length-dependent logarithmic tolerance function therefore offers a compact way to reconcile (i) diminishing marginal perceptual impact of individual errors with (ii) increasing cumulative processing costs as disruptions accumulate across longer texts. This theoretical synthesis motivates the non-linear scoring model introduced in Section~\ref{sec:NLS}.

\section{Proposed Non-Linear Scoring Model}
\label{sec:NLS}

We adopt the terminology introduced in our primary study \cite{GladkoffEtAl2024}, which distinguishes three classes of scoring approaches for MQM-style analytic Translation Quality Evaluation (TQE):

\begin{mdframed}
\begin{itemize}
    \item \textbf{Raw score:}
    \[
    100 - \alpha\,\mathrm{APT}/\mathrm{EWC},
    \]
    where \(\alpha\) is a unit scaling factor (e.g., \(\alpha=1000\) for ``per 1{,}000 words''). This formulation reflects the sample’s observed penalty \emph{rate} but is not anchored to an institutional tolerance threshold and does not yield a calibrated quality scale.

    \item \textbf{Calibrated linear model:}
    a scoring model that maps an institution’s acceptability tolerance (pass/fail threshold) onto a normalized 0--100 scale, enabling interpretability and comparability within a fixed reference range.

    \item \textbf{Calibrated non-linear model:}
    a scoring model that replaces proportional scaling with a length-dependent tolerance function, modeled here as a logarithmic curve. Relative to a linear rule anchored at a reference size, the non-linear model is more permissive for short samples and more stringent for long samples; total tolerance still increases with length, but at a diminishing rate.
\end{itemize}
\end{mdframed}

Given the empirical results in Section~\ref{sec:empirical} and the theoretical motivation in Section~\ref{sec:physhophysical}, we do not argue for eliminating linear scoring entirely. Rather, we propose a principled regime of applicability: linear scoring can remain useful near its reference calibration size, while a non-linear tolerance function is required for robust length generalization.

In particular, even within the meso-range (up to approximately 3{,}000 words), linear extrapolation can introduce substantial distortion when applied outside its reference vicinity. For example, as shown in Section~\ref{subsec:fidelity-interval}, a linear model anchored at a 1{,}000-word reference point can exceed a 20\% relative deviation when applied to a 2{,}000-word sample. Accordingly, we recommend using the logarithmic model whenever evaluation sample sizes are expected to vary by more than approximately \(20\%\) from the reference word count, i.e., outside the $\pm 20\%$ fidelity interval defined in Section~\ref{subsec:fidelity-interval}.

Raw scoring and calibrated linear scoring remain viable within restricted settings, provided that evaluation samples consistently remain near the calibration length.\footnote{%
For raw scoring, ``calibration'' amounts to selecting a pass/fail cutoff on the penalty-rate scale. Let \(r_{\text{thr}}\) denote the maximum acceptable penalty rate (e.g., 5 points per 1{,}000 words). A sample with penalty total \(\mathrm{APT}\) and word count \(\mathrm{EWC}\) passes iff \(\mathrm{APT}/\mathrm{EWC} \le r_{\text{thr}}\), equivalently \(\mathrm{APT} \le r_{\text{thr}}\cdot \mathrm{EWC}\). No mapping to a normalized 0--100 scale is involved. Some implementations rescale the penalty rate into a 0--100 ``raw score'' via \(100 - 1000\cdot\mathrm{APT}/\mathrm{EWC}\); however, this inherits the same limitations of proportional length scaling. To obtain scores that remain comparable across varying evaluation sizes, a calibrated model is required (linear near the reference size, non-linear over wider ranges).}

This section introduces the mathematical form of the proposed logarithmic tolerance model, describes its anchoring and calibration procedures, and characterizes the regime in which linear approximation remains sufficiently accurate. By extending MQM scoring with a length-dependent tolerance function, the proposed approach enables analytic evaluation to better align with document-level expert judgments under variable-length sampling conditions.

\subsection{Limitations of Linear Extrapolation}

Linear scoring models assume that acceptable error counts increase proportionally with sample length. For example, if an institution accepts 5 minor errors in a 1{,}000-word sample, linear scaling permits 10 minor errors in 2{,}000 words and 15 in 3{,}000 words. However, as shown in Sections~\ref{sec:empirical} and~\ref{sec:physhophysical}, this assumption is not consistent with elicited expert tolerances across variable-length samples.

In practice, institutional guidance may appear linear when expressed informally (e.g., ``approximately one serious error per page''), yet observed acceptance judgments are typically sublinear. When sample size increases to multiple pages, evaluators do not generally accept a strictly proportional increase in severe errors; instead, tolerance per unit length decreases with text length. This indicates that proportional extrapolation can systematically overestimate acceptable error totals for longer samples and underestimate the impact of errors in shorter samples.

\subsection{Proposed Mathematical Model of Acceptable Errors}

We model the maximum acceptable number of errors \(E(x)\) as a function of sample size \(x\) (measured in words) using the following logarithmic tolerance function:
\begin{equation}
E(x) \;=\; a\,\ln(1 + b\,x),
\qquad a>0,\; b>0.
\label{eq:log-model}
\end{equation}
Throughout, \(\ln\) denotes the natural logarithm. The function satisfies \(E(0)=0\), increases rapidly for short samples, and flattens gradually as sample length increases, matching the concave tolerance trend observed empirically.

If sample size is expressed in pages with \(W\) words per page, the same curve is obtained by replacing \(b\) with \(Wb\), while \(a\) remains unchanged. This unit invariance ensures consistency between page-based visualizations and word-based formulations.

We adopt the intercept-free specification \(E(x)=a\ln(1+bx)\) to enforce the natural boundary condition \(E(0)=0\) and to provide explicit control of curvature via parameter \(b\). This differs from the ``logarithmic trendline'' form \(c + k\ln x\) commonly used in spreadsheet software, which introduces a free intercept term and does not pass through the origin. As a result, the spreadsheet form is not directly anchorable to tolerance points in the same manner.

Other scoring models may be developed for specific applications. Importantly, however, severity weighting within a fixed sample is distinct from the length-dependent non-linearity considered here. Even an exponential ladder of severity multipliers affects how errors are aggregated \emph{within} a sample, but it does not, by itself, induce a non-linear relationship between allowable error totals and sample size.

For example, ATA certification scoring is an analytic, points-off scheme tailored to short, fixed-length exam translations, with steep severity scaling but no tolerance curve \(E(x)\) varying with length. Similarly, SAE~J2450 assigns fixed category weights and sums penalties additively within a sample \cite{saeJ2450}. Such frameworks introduce non-uniformity across severities, but not the sample-size dependent tolerance function required to keep pass/fail criteria comparable across widely varying evaluation lengths.

\subsection{Anchoring and Calibration Procedures}
\label{sec:calibration}

For operational deployment, we distinguish two adaptation steps:

\begin{itemize}
    \item \textbf{Anchoring:} specifying an acceptable error limit at a reference sample size, expressed in severity-weighted penalty points (e.g., ``no more than 5 points per 1{,}000 words'').
    \item \textbf{Calibration:} mapping the observed penalty total for a sample into a normalized 0--100 quality scale, supporting interpretability in dashboards and reporting.
\end{itemize}

\textbf{Two-point feasibility.}
Let \((x_0,E_0)\) and \((x_1,E_1)\) be two tolerance points with \(x_0\neq x_1\) and \(E_0,E_1>0\). Define
\[
r:=\frac{E_1}{E_0},\qquad \rho:=\frac{x_1}{x_0},\qquad g(b):=\frac{\ln(1+b x_1)}{\ln(1+b x_0)}.
\]
For \(b>0\), \(g\) is continuous and strictly monotone, with \(\lim_{b\downarrow 0}g(b)=\rho\) and \(\lim_{b\uparrow\infty}g(b)=1\). Hence a unique \(b^\star>0\) exists if and only if
\[
\boxed{\ \min\{1,\rho\}\;<\;r\;<\;\max\{1,\rho\}\ }.
\]
Intuitively, the longer sample must permit more errors, but sublinearly, relative to its length increase.

\textbf{Two-point calibration (numerical).}
Let \(r=E_1/E_0\) and define \(f(b)=\ln(1+b x_1)-r\,\ln(1+b x_0)\). Under the feasibility condition above, \(f\) has exactly one root \(b^\star>0\). The parameter \(b^\star\) can be obtained with a robust one-dimensional root-finding method (e.g., bisection or Brent), after which
\[
a^\star=\frac{E_0}{\ln(1+b^\star x_0)}.
\]

See~\nameref{app:two-point-cal} for a one-dimensional root-finding implementation.

\medskip
\noindent\textit{\textbf{Illustrative values}.}\\
For the example anchor points \((x_0,E_0)=(1000,5)\) and \((x_1,E_1)=(250,2)\), solving \(f(b)=0\) yields
\(b^\star \approx 2.880\times 10^{-3}\) and \(a^\star \approx 3.688\), giving
\[
E(x) \;=\; 3.688\,\ln\!\bigl(1 + 0.00288\,x\bigr).
\]

\medskip
\noindent\textit{Fallback for multiple points.}
If more than two tolerance points \(\{(x_i,E_i)\}_{i=1}^n\) are available, estimate \((a,b)\) via constrained least squares:
\[
  \min_{a>0,\;b>0}\ \sum_{i=1}^{n}\Bigl[E_i - a\,\ln\!\bigl(1+b\,x_i\bigr)\Bigr]^2.
\]
Given any \(b_0>0\), a fast one-dimensional search over \(b\) with closed-form
\[
a(b)=\frac{\sum_i E_i\,\ln(1+bx_i)}{\sum_i \ln^2(1+bx_i)}
\]
provides an efficient initialization (followed by optional joint refinement of \((a,b)\)).

For small \(b\), \(\ln(1+bx)\approx bx\), implying that the model is locally linear near the reference size. This explains why calibrated linear scoring can be adequate only in a neighborhood around its anchor point.

\begin{figure}[ht]
    \centering
    \includegraphics[width=1\linewidth]{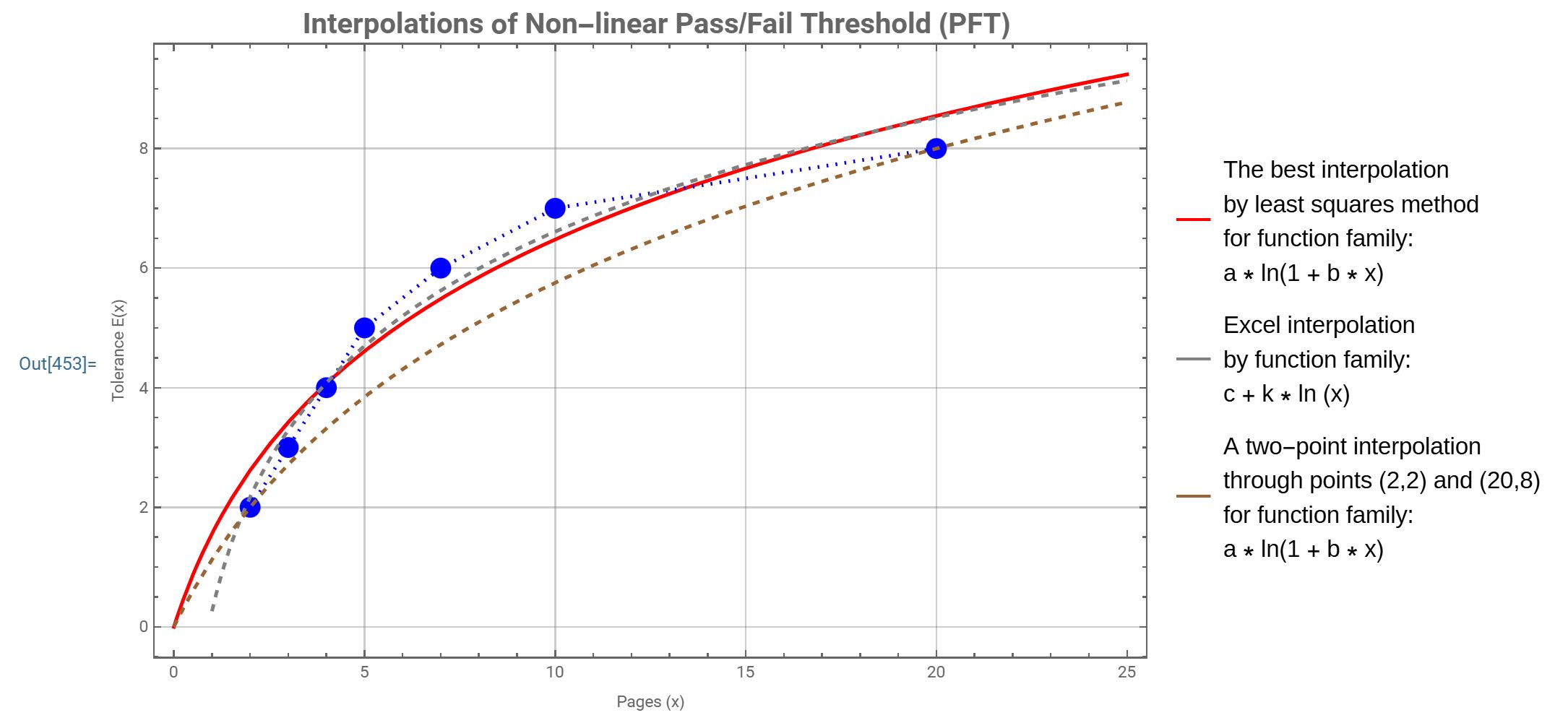}
    \caption{\textbf{Three interpolations.} The spreadsheet trendline \(c+k\ln x\) is shown for comparison only; calibration uses \(E(x)=a\ln(1+bx)\), which passes through the origin and is anchorable to tolerance points.}
    \label{fig:threeinterpolationtypes}
\end{figure}

\subsection{Near-Reference Fidelity Interval for Linear Approximation ($\pm20\%$)}
\label{subsec:fidelity-interval}

In practice, a linear rule anchored at a reference size \(x_{\mathrm{ref}}\) is often used:
\[
E_{\mathrm{lin}}(x)
\;=\;
E_{\log}(x_{\mathrm{ref}})\,\frac{x}{x_{\mathrm{ref}}},
\qquad
E_{\log}(x)=a\,\ln\!\bigl(1+bx\bigr).
\]
To characterize where this linear approximation remains acceptable, we define the \emph{$\pm20\%$ fidelity interval} around \(x_{\mathrm{ref}}\) as the set of \(x>0\) such that
\[
\left|\frac{E_{\mathrm{lin}}(x)}{E_{\log}(x)}-1\right|\le 0.20.
\]

Writing \(\varepsilon=0.20\) and \(\alpha=\dfrac{\ln(1+b\,x_{\mathrm{ref}})}{b\,x_{\mathrm{ref}}}\), the boundary points admit a closed form using the Lambert \(W\) function:
\[
x_{\pm}
\;=\;
\frac{-\,\dfrac{1\pm \varepsilon}{\alpha}\,
W_{-1}\!\!\left(\,-\frac{\alpha}{1\pm \varepsilon}\,
e^{-\alpha/(1\pm \varepsilon)}\right)-1}{\,b\,},
\]
where \(x_{-}\) corresponds to the lower boundary and \(x_{+}\) to the upper boundary.\footnote{The branch \(W_{-1}\) yields the solution near \(x_{\mathrm{ref}}\).}

Using the illustrative calibration in Section~\ref{sec:calibration} (\(a=3.688,\,b=0.00288\)), the resulting fidelity intervals are:
\begin{itemize}
  \item \(\boldsymbol{x_{\mathrm{ref}}=1000}\): \(x\in[579,\;1460]\) words.
  \item \(\boldsymbol{x_{\mathrm{ref}}=2000}\): \(x\in[1307,\;2747]\) words.
\end{itemize}

These intervals quantify the limited range over which a single anchored linear slope remains within 20\% relative deviation. Outside this range, proportional extrapolation becomes increasingly inaccurate, motivating use of the logarithmic model or re-anchoring the linear model to a more appropriate reference size.

\subsection{Visual comparison across sample-size regimes}

\begin{figure}[ht]
    \centering
    \includegraphics[width=1\linewidth]{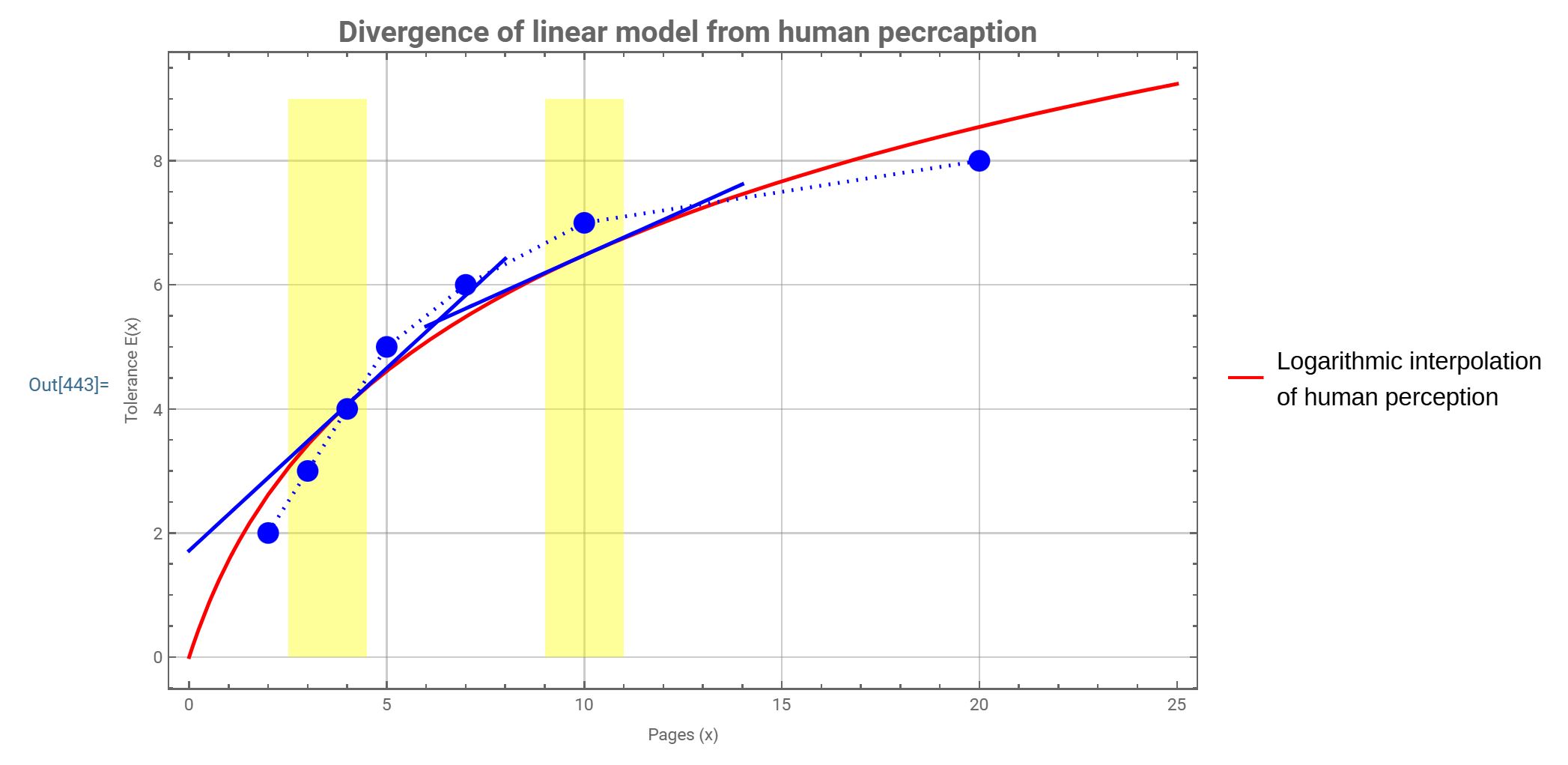}
    \caption{\textbf{Linear model vs.\ non-linear tolerance.} The shaded bands indicate the $\pm20\%$ fidelity zone around the linear anchor; outside these bands the linear rule under- or over-estimates tolerance.}
    \label{fig:lineardivergence}
\end{figure}

Figure~\ref{fig:lineardivergence} illustrates that once the slope is fixed by a single anchor point, a linear rule diverges substantially from the concave tolerance pattern implied by human judgments at both smaller and larger evaluation sizes. In contrast, the logarithmic model remains consistent with the length-dependent tolerance trend across the range of sample sizes considered.

\begin{figure}[ht]
    \centering
    \includegraphics[width=0.85\linewidth]{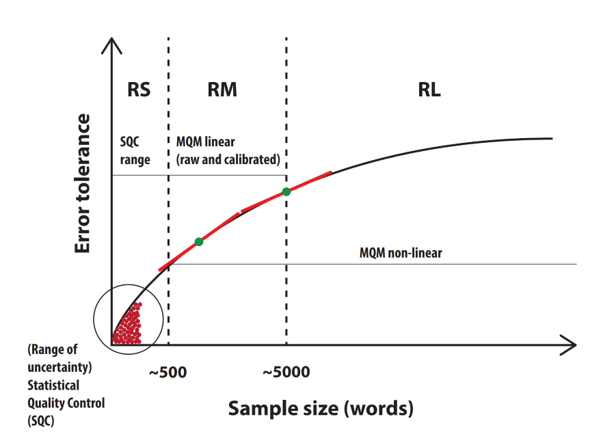}
    \caption{\textbf{Applicability across evaluation sizes.} SQC (micro, $\lesssim 250$ words), calibrated linear scoring (meso; reliable only near anchor), and non-linear scoring (macro). Y-axis: tolerance in penalty points.}
    \label{fig:scoring_ranges}
\end{figure}

Finally, Table~\ref{tab:guidance} summarizes operational guidance for selecting scoring models under varying sampling conditions.

\begin{table}[htbp]
\centering
\begin{tabular}{@{}p{0.55\linewidth}p{0.35\linewidth}@{}}
\toprule
Scenario & Recommended model \\
\midrule
EWC varies within $\pm 20\%$ of the anchor & Calibrated linear model (re-anchor if the range shifts) \\
EWC routinely outside the $\pm 20\%$ band & Use the non-linear logarithmic model directly \\
Very short samples ($\lesssim 250$ words) & Statistical Quality Control (SQC) \\
\bottomrule
\end{tabular}
\caption{\textbf{Operational guidance for model selection.}}
\label{tab:guidance}
\end{table}

\section{Practical Implications}
\label{sec:practical}

If error tolerance scales non-linearly with sample length---as suggested by the elicitation results in Section~\ref{sec:empirical} and the theoretical considerations in Section~\ref{sec:physhophysical}---then adopting a logarithmic tolerance function has practical implications for translation quality evaluation procedures, quality management policies, and supporting software tooling. In particular, replacing proportional length normalization with a calibrated, length-dependent tolerance model can reduce systematic length-induced bias and improve the consistency of acceptability decisions across samples of varying size.

A length-dependent scoring model may also be relevant for emerging AI-assisted document-level evaluation workflows. Contemporary systems can provide holistic judgments, but their outputs are often qualitative and may vary in completeness or granularity across documents. A calibrated non-linear scoring model offers a quantitative reference function that could be used to normalize or contextualize such judgments against a stable, institution-specific 0--100 scoring scale (e.g., via a user-specific PT/MSV mapping), thereby facilitating comparability across evaluation settings.

\subsection{Operational Guidelines for Quality Managers}

A direct operational benefit of the proposed model is support for consistent tolerance setting across variable evaluation sample sizes. Quality managers can use the calibrated logarithmic function \(E(x)\) to define explicit acceptability thresholds as a function of evaluation word count, and to adapt tolerances across content types or use cases without relying on ad hoc proportional extrapolation.

For example, a quality policy can be expressed as follows: \emph{``We anchor tolerance at 2{,}000 words with a maximum of 7 minor-error equivalents. Linear approximation is acceptable within the fidelity interval (approximately 1{,}500--3{,}000 words in this configuration). Outside this range, tolerance thresholds should be computed using the logarithmic model.''}

This approach enables stakeholders to move beyond a single fixed-length scorecard and instead apply acceptance thresholds that reflect length-dependent judgment behavior. The resulting evaluation outcomes are more interpretable and more equitable across texts of different lengths, particularly when quality monitoring requires comparison across heterogeneous content types and evaluation sample sizes.

\subsection{Tooling Implications}

From a tooling perspective, the proposed model is straightforward to integrate into existing MQM-based evaluation pipelines. Most production systems already compute the severity-weighted penalty total (\textsc{APT}) and the evaluated word count (\textsc{EWC}); therefore, incorporating non-linear length dependence requires only replacing a constant tolerance threshold with a length-dependent tolerance function \(E(x)\). Concretely, the acceptability condition becomes
\[
\mathrm{APT} \le E(\mathrm{EWC}),
\]
rather than \(\mathrm{APT} \le r_{\text{thr}}\cdot \mathrm{EWC}\) under proportional scaling.

Because the core annotation workflow, error typology, and penalty aggregation remain unchanged, the model can be deployed without modifying MQM schemas, annotation tools, or reviewer training protocols. The primary implementation changes occur at the scoring layer and in reporting interfaces (e.g., dashboards and QA summaries), which must display tolerances and pass/fail thresholds that vary with sample length. In addition, calibration parameters \((a,b)\) can be stored per institution, content type, or language pair, enabling consistent application within a unified scoring framework while preserving user-specific acceptance standards.


\paragraph{Implementation sketch.}
Algorithm~\ref{alg:nls-implementation} summarizes a minimal integration of length-dependent tolerance into a typical MQM scoring pipeline.

\begin{algorithm}[ht]
\caption{Length-dependent tolerance scoring (non-linear MQM)}
\label{alg:nls-implementation}
\begin{algorithmic}[1]
\Require Annotated errors with severities; evaluated word count \(\mathrm{EWC}\); calibrated parameters \((a,b)\); score mapping function \(\mathrm{Score}(\cdot)\)
\Ensure Pass/Fail decision; normalized 0--100 score

\State Compute severity-weighted penalty total \(\mathrm{APT}\) from annotations
\State Compute tolerance threshold \(T \gets a\ln(1 + b\,\mathrm{EWC})\)
\State \(\mathrm{Pass} \gets (\mathrm{APT} \le T)\)

\Statex
\Comment{Optional: normalized reporting scale}
\State Compute deviation ratio \(d \gets \mathrm{APT}/T\)
\State Compute normalized score \(s \gets \mathrm{Score}(d)\)
\State \Return \(\mathrm{Pass},\, s\)
\end{algorithmic}
\end{algorithm}






\section{Building a Non-Linear Scorecard}
\label{sec:building}

This section provides a practical procedure for constructing and deploying a non-linear MQM scorecard, covering (i) model fitting from tolerance points, (ii) conversion from tolerance to a calibrated score on a 0--100 scale, and (iii) scorecard field computation for operational reporting. We further discuss implications for CAT tools and LQA automation, the benefits of a unified calibrated quality scale, applications to AI-generated content, and document-level evaluation considerations.\footnote{The designations and symbols in this section follow the variable naming used in the ASTM working item WK46396 (``MQM~2.0: Analytic Translation Quality Evaluation''), which describes MQM scoring models and scoring mechanics in detail. \url{http://www.astm.org/workitem-wk46396}.}

\subsection{Model and least-squares fit}

We model pass/fail tolerance as a function of evaluation sample size \(x\) (in words) using the logarithmic form
\begin{equation}
  E(x) \;=\; a \,\ln\!\bigl(1+b\,x\bigr), \qquad a>0,\;b>0.
  \label{eq:logmodel}
\end{equation}
Given tolerance points \(\{(x_i,E_i)\}_{i=1}^n\), the parameters \((a,b)\) can be estimated by constrained least squares:
\[
  \min_{a>0,\;b>0}\;\sum_{i=1}^n \bigl[E_i - a\,\ln(1+b\,x_i)\bigr]^2.
\]
For any fixed \(b>0\), the optimal \(a\) admits the closed-form expression
\begin{equation}
  a(b)\;=\;\frac{\sum_i E_i\,\ln(1+b\,x_i)}{\sum_i \ln^2(1+b\,x_i)} .
\end{equation}
Substituting \(a(b)\) yields a one-dimensional objective
\[
S(b)=\sum_i\!\bigl[E_i-a(b)\ln(1+b\,x_i)\bigr]^2,
\]
which can be minimized over \(b>0\), after which \(\hat a := a(\hat b)\). Two-point anchoring, feasibility conditions, and the full least-squares procedure are detailed in Section~\ref{sec:calibration} and Appendix~A--B.

If the model is fitted using pages with \(W\) words per page and later applied using word counts, only \(b\) rescales: \(b_{\text{words}} = b_{\text{pages}}/W\), while \(a\) remains unchanged.

\subsection{From tolerance to a calibrated score}

For an evaluation sample of size \(x=\mathrm{EWC}\) (words), the tolerance threshold at the passing boundary is computed as
\begin{equation}
  E_{\text{allowed}}(x) \;=\; \hat a\,\ln\!\bigl(1+\hat b\,x\bigr).
\end{equation}
Let \(\mathrm{APT}\) denote the observed absolute penalty total from the MQM annotation table. We define the quality fraction as
\begin{equation}
  \mathrm{QF}(x) := 1 - \frac{\mathrm{APT}}{E_{\text{allowed}}(x)} .
\end{equation}
The quality fraction is then mapped onto a calibrated score using the Maximum Score Value (MSV), Passing Threshold (PT), and Defined Passing Interval \(\mathrm{DPI}=\mathrm{MSV}-\mathrm{PT}\):
\begin{equation}
  \mathrm{OS}(x) := \mathrm{PT} + \mathrm{DPI}\,\mathrm{QF}(x),
  \qquad
  \mathrm{OS}_{\text{disp}} := \min\{\mathrm{MSV},\,\max\{0,\mathrm{OS}(x)\}\}.
  \label{eq:calibrated}
\end{equation}
The evaluation outcome is \textbf{Pass} if \(\mathrm{APT}\le E_{\text{allowed}}(x)\), and \textbf{Fail} otherwise.

\textbf{Monotonicity and clipping.}
Since \(\mathrm{OS}(x)\) is affine in \(\mathrm{APT}\), it decreases monotonically as \(\mathrm{APT}\) increases. The displayed score \(\mathrm{OS}_{\text{disp}}\) is clipped for reporting convenience and does not affect pass/fail decisions, which are determined solely by the inequality \(\mathrm{APT}\le E_{\text{allowed}}(x)\).

\subsection{Scorecard mapping}

Table~\ref{tab:scorecard-mapping} summarizes the computations required to populate an operational scorecard using the evaluation word count (EWC), fitted coefficients \(\hat a,\hat b\), and severity-weighted penalty total (APT). Importantly, the MQM annotation process and error-type/severity schema remain unchanged; only the tolerance threshold becomes length-dependent via Equation~\eqref{eq:logmodel}.

\begin{table}[!htbp]
\centering
\begin{tabular}{@{}ll@{}}
\toprule
\textbf{Name} & \textbf{Excel-style computation} \\
\midrule
Evaluation size \(x\) (EWC) & input \\
Model coefficients \(\hat a, \hat b\) & calibrated from tolerance points (Section~\ref{sec:calibration}) \\
Allowed penalty \(E_{\mathrm{allowed}}(x)\) & \texttt{= a*LN(1 + b*EWC)} \\
Quality fraction \(QF(x)\) & \(\displaystyle 1 - \frac{\mathrm{APT}}{E_{\mathrm{allowed}}}\) \\
Calibrated score \(\mathrm{OS}\) & \texttt{= PT + (MSV-PT) * QF} \\
Displayed score \(\mathrm{OS}_{\mathrm{disp}}\) & \texttt{= MIN(MSV, MAX(0, OS))} \\
Decision margin \(DM\) & \(E_{\mathrm{allowed}} - \mathrm{APT}\) \\
Pass/Fail & \texttt{= IF(APT <= Eallowed,"PASS","FAIL")} \\
\bottomrule
\end{tabular}
\caption{\textbf{Scorecard mapping: variables and Excel-style computations.}}
\label{tab:scorecard-mapping}
\end{table}

\begin{figure}[ht]
    \centering
    \includegraphics[width=0.85\linewidth]{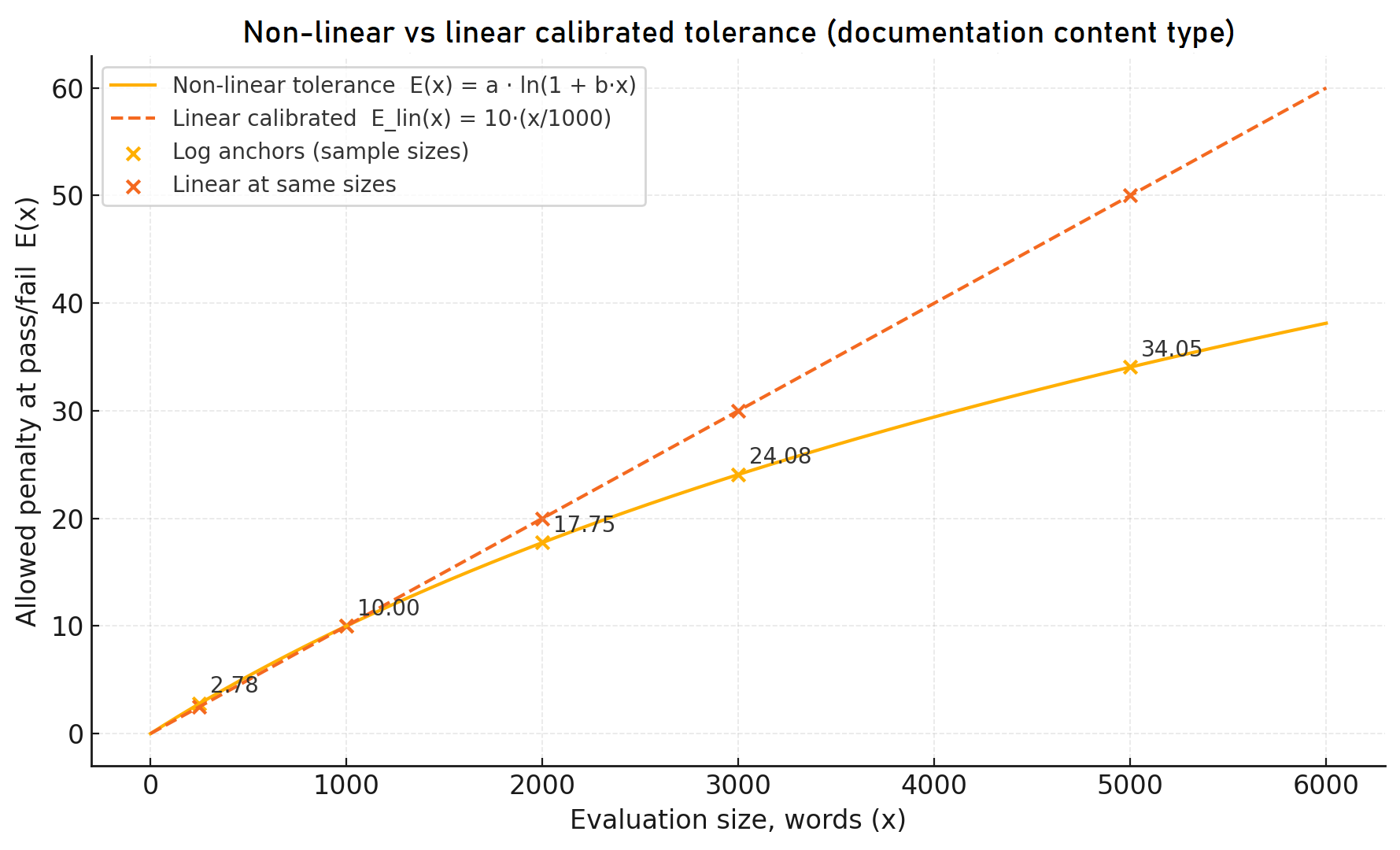}
    \caption{\textbf{Example non-linear tolerance curve} (solid) compared with an anchored linear rule (dashed). The logarithmic curve is calibrated from tolerance points and represents institutional tolerance as a function of sample size; the linear curve is shown for reference.}
    \label{fig:SAP-curve}
\end{figure}

Figure~\ref{fig:SAP-curve} illustrates a real-world non-linear tolerance curve calibrated from institutional anchor points, alongside the corresponding anchored linear rule for comparison.

\textbf{Same-curvature re-anchoring (optional).}
To impose a new anchor value \(E(x_{\text{ref}})=E_0\) while preserving curvature, fix \(b\) and solve
\[
  a \;=\; \frac{E_0}{\ln(1+b\,x_{\text{ref}})} .
\]

\textbf{Micro-range caution.}
For sample sizes below approximately 250 words, deterministic tolerance curves are statistically unstable, and Statistical Quality Control (SQC) procedures are recommended instead. Intuitively, if the underlying minor-error rate is \(p\) per word and only \(n\) words are observed, the expected count is \(np\) with standard deviation \(\sqrt{np(1-p)}\approx\sqrt{np}\), so relative uncertainty scales as \(1/\sqrt{np}\) and becomes large when \(np \approx 1\).

\begin{mdframed}
\textbf{Worked example (documentation content).}
Calibrated tolerance: \(E(x)=3.688\,\ln(1+0.00288\,x)\).
For \(x=3000\) words, \(E_{\mathrm{allowed}}(x)\approx 8.357\) points.
If \(\mathrm{APT}=7\), then \(QF\approx 0.162\).
With \(\mathrm{PT}=80\), \(\mathrm{MSV}=100\) (\(\mathrm{DPI}=20\)), \(\mathrm{OS}\approx 83.25\) and \(DM\approx 1.36\) (\textbf{Pass}).
If \(\mathrm{APT}=9\), then \(QF\approx -0.077\), \(\mathrm{OS}\approx 78.46\) and \(DM\approx -0.64\) (\textbf{Fail}).
\end{mdframed}

\subsection{Implications for CAT tools and LQA automation}

Adopting a logarithmic tolerance function has direct implications for software systems implementing LQA workflows, including CAT tools with integrated quality modules and standalone QA platforms. Under proportional scaling, thresholds are defined using a constant error density relative to a fixed reference length; under the proposed model, tolerance must be computed dynamically as a function of the observed evaluation word count.

In practice, this requires tooling to:
\begin{itemize}
    \item replace static pass/fail thresholds with a length-dependent tolerance function \(E(x)\);
    \item visualize quality relative to a curved tolerance baseline rather than a proportional reference line; and
    \item report decision margins (e.g., \(DM=E_{\mathrm{allowed}}-\mathrm{APT}\)) to communicate proximity to the pass/fail boundary.
\end{itemize}

These adaptations are particularly relevant in workflows involving heterogeneous sample sizes, such as machine translation, post-editing, and AI-assisted content generation.

\subsection{Toward a unified calibrated quality scale}

A persistent challenge in analytic TQE is maintaining interpretability and comparability across content types, language pairs, and evaluation sample lengths. Calibrated scoring improves over raw penalty rates by mapping results to a stable reporting scale; however, proportional length scaling can still distort comparability when applied outside its effective range.

By incorporating length-dependent tolerance via \(E(x)\), non-linear calibrated scoring supports a unified, institution-specific 0--100 scale for heterogeneous evaluation sizes. This improves comparability across jobs and enables more reliable aggregation for analytics such as trend monitoring, supplier benchmarking, and root-cause analysis.

As a concrete example, \cite{GladkoffEtAl2024} reports a case in which a 5{,}000-word English--German marketing translation containing 23 minor errors passed under linear scaling but was rejected by expert reviewers. The logarithmic model yields a substantially lower tolerance threshold in this regime (16 minor-error equivalents), consistent with the human accept/reject decision.

\subsection{Applications beyond human translation}

Although the model is motivated by human translation evaluation practices, the same length-dependent tolerance considerations apply to AI-generated text. Large language model (LLM) outputs (e.g., summaries, answers, or translations) are increasingly evaluated using procedures adapted from TQE and LQA \cite{han2024neural}. Non-linear tolerance modeling provides a more stable baseline for such evaluation by reducing over-penalization of short segments and preventing under-penalization of accumulated minor issues in long outputs.

\subsection{Document-level evaluation and future automation}
\label{sec:practical_context}

Document-level quality assessment requires integrating local defects with discourse-level properties such as cohesion, register consistency, and terminology management. Small samples can provide rapid quality signals, but they provide only partial visibility into document-level phenomena and may lead to unstable conclusions when length effects are ignored.

The MQM-based non-linear calibrated model addresses this by maintaining linear behavior near its reference regime while extending smoothly to longer samples via a single tolerance function. Provided the evaluation sample is sufficiently large to yield statistically stable estimates (as recommended by the Multi-Range framework \cite{GladkoffEtAl2024}), the resulting scores remain consistent with document-level expert judgments.

Finally, although current automated systems do not yet perform complete MQM-style analytic evaluation over long contexts, future evaluation models may increasingly combine local error detection with document-level assessment. In such settings, a length-dependent tolerance function provides a principled backbone for ensuring that automated scoring remains comparable and perceptually aligned across a wide range of evaluation lengths.

\subsection{Summary}
\label{subsec:sumary}

In practical terms, transitioning to a non-linear tolerance model enables:
\begin{itemize}
    \item improved consistency of acceptability decisions across variable-length evaluation samples;
    \item more interpretable and equitable calibrated scores on a unified reporting scale;
    \item simpler integration into CAT and LQA tooling via dynamic tolerance computation; and
    \item better alignment with document-level evaluation requirements for both human and AI-generated content.
\end{itemize}

Overall, the proposed approach operationalizes empirically supported non-linear tolerance scaling in a form that is simple to calibrate, straightforward to implement, and compatible with existing MQM-based workflows.

\section{Limitations and Future Work}
\label{sec:limitation}

This section qualifies the practical benefits summarized in Section~\ref{subsec:sumary} and identifies directions for further validation and extension. Although the proposed non-linear tolerance model is supported by both expert elicitation and theoretical motivation, several limitations constrain the scope of the current evidence base.

\subsection{Sample size and elicitation design}

The empirical evidence in Section~\ref{sec:empirical} is based on structured elicitation with Quality Managers from three large institutional translation programs. While these organizations operate mature, high-volume multilingual workflows, the number of participating institutions (and respondents) remains limited. Moreover, tolerance judgments were elicited rather than measured in blinded controlled experiments, which may introduce subjectivity and potential response bias.

Future work should expand the sample to include a broader range of stakeholders---e.g., LSPs, in-house localization teams, and public-sector translation units---across domains and language pairs. Controlled studies could further strengthen validity by comparing (i)~holistic accept/reject decisions and (ii)~analytic MQM-based scoring outcomes under alternative tolerance functions, enabling direct assessment of predictive alignment and decision consistency.

\subsection{Content-type and domain effects}

Quality expectations vary by content type (e.g., marketing, legal, technical documentation, UI strings). The proposed model assumes a common functional form \(E(x)=a\ln(1+bx)\), with content- or user-specific variation captured by the parameters \(a\) (scale) and \(b\) (curvature). Our operational observations suggest that a single logarithmic form can perform robustly across heterogeneous content types, but broader field evidence is required.

An important extension is to test whether particular domains exhibit systematically different curvature profiles (e.g., steeper for high-risk instructional content or flatter for low-stakes internal communication), and whether additional saturation occurs for very large evaluation sizes (e.g., book- or manual-length contexts). If such effects are observed, alternative sublinear families (e.g., log--log, logistic saturation, or piecewise concave forms) may be warranted.

\subsection{Severity weighting and mixed error distributions}

This paper models tolerance in severity-weighted penalty points, where one point is typically interpreted as one minor-error equivalent. In operational settings, evaluations include mixed severities (minor/major/critical) with fixed multipliers and may also include error-type weights. Although the logarithmic tolerance curve captures minor-error tolerance well---minor errors being the most frequent and thus statistically most stable---rarer severities may require discrete (integer) decision rules or severity-specific tolerance sequences.

Future work should evaluate how tolerance behaves for major and critical errors under variable-length sampling. This includes establishing principled rounding and decision rules, testing whether severity-specific tolerance sequences remain approximately sublinear, and assessing whether severity multipliers interact with length-dependent tolerance in ways that call for multi-parameter extensions.

\subsection{Inter-rater reliability}

A key motivation for length-aware tolerance is to reduce systematic disagreement caused by applying proportional extrapolation outside its effective reference region. The proposed model is expected to improve inter-rater reliability (IRR) by providing more consistent pass/fail thresholds across varying sample sizes. However, the present work does not yet quantify IRR improvements under controlled rater studies.

Future work should measure IRR (e.g., using Krippendorff’s \(\alpha\), Cohen’s \(\kappa\), or intra-class correlation, depending on task design) under both linear and non-linear scoring regimes. Such studies could test whether a length-dependent tolerance model reduces variance in borderline cases and improves agreement across evaluators with different experience levels.

\subsection{Extension to AI evaluation benchmarks}

We hypothesize that similar length-dependent tolerance effects may apply to the evaluation of AI-generated text (including document translation, summarization, and dialogue). This claim remains to be validated empirically. Benchmarking settings---including shared-task evaluations in MT such as WMT \cite{kocmi-etal-2025-findings}---often rely on reference-based metrics or length-normalized error densities; incorporating calibrated non-linear normalization may improve comparability across variable-length outputs and strengthen alignment with human preferences in document-level evaluation.

Future work should test the proposed model as a normalization layer for paragraph- and document-level evaluation in both MT and broader LLM generation tasks, including analysis of robustness across domains and prompt conditions. A longer-term direction is to investigate whether calibrated tolerance functions can serve as quantitative scaffolding for AI-assisted document-level quality assessment systems that currently produce holistic but weakly calibrated judgments.

\section{Conclusion}
\label{sec:conclude}

This paper addressed the length-dependence of acceptability thresholds in MQM-based analytic Translation Quality Evaluation (TQE). While proportional (linear) error scaling is widely used in practice, our results indicate that it can produce systematic bias when evaluation sample sizes deviate from the reference word count.

First, we provided empirical evidence from three large institutional translation programs showing that tolerable error counts increase with sample length \emph{sublinearly}, yielding concave tolerance profiles consistent with logarithmic growth. Second, we linked this behavior to established theoretical accounts from psychophysics (compressive perceptual scaling) and Cognitive Load Theory (accumulated processing disruption), which together motivate non-linear integration of errors over increasing text length. Third, we proposed a calibrated two-parameter tolerance function \(E(x)=a\ln(1+bx)\) and showed how it can be anchored from tolerance points using a simple one-dimensional calibration procedure. Fourth, we demonstrated that the model can be integrated into existing MQM pipelines with minimal change: the MQM annotation process and penalty computation remain unchanged, while a constant length-normalized tolerance is replaced by a length-dependent threshold function.

In addition, we derived a near-reference fidelity interval that characterizes when a linear approximation remains sufficiently accurate (e.g., within \(\pm20\%\)), providing a practical guideline for model selection in operational deployments. Finally, we reiterate that for very short samples (approximately \(\lesssim250\) words), deterministic tolerance curves are statistically unstable, and risk-based Statistical Quality Control (SQC) remains the appropriate framework for decision-making under high sampling uncertainty \cite{GladkoffEtAl2024}.

\section{acknowledgments}
The authors gratefully acknowledge the MQM Council and the ASTM WK46396 Working Group
(MQM 2.0: Analytic Translation Quality Evaluation) for providing a collaborative forum in which
earlier versions of the non-linear scoring model were discussed, refined, and incorporated into the
broader MQM framework. In particular, the authors would like to thank Ingemar Strandvik (MQM
Council) and Angelika Vasaa, Head of the Quality Coordination Unit at the European Parliament,
for their insightful comments, institutional perspective, and many valuable discussions that helped
shape the ideas presented in this paper. 
Views and opinions expressed are however their own of the authors and acknowledged contributors, and do not necessarily reflect those of the European Union or Parliament.
Neither the European Union nor the
Parliament's authority can be held responsible for them.\\
The authors thank the Quality Managers of the participating organizations (referred to in this
document as users C1--C3) for their generous contributions of time, insight, and real-world data.
Their candid responses and openness to critical reflection were essential to this research. Gratitude is
also expressed to the broader translation quality evaluation community for ongoing discussions that
continue to refine our shared understanding of what quality means in multilingual communication.

\begin{appendices}

\section{Institution User Quotations on  Evaluation}\label{Appendix_users_quote}

In points‑off situations, the linear rule makes long samples appear better than experts judge them to be by raising the acceptance threshold in direct proportion to length.

A recent quote from a large institutional user illustrates this point:

\begin{quote}
``Once we started using our current methodology in 2020, we still asked the evaluators to indicate the cases where their actual feeling was different from what the score gave them. We very quickly realized that the main issue was that with very short samples the scoring was overly harsh and with very long samples it was too lenient. The reason for this is that when we evaluate holistically, the perception is not congruent with our scoring formula. For example, we might feel that if a translation sample is about one page, a single major mistranslation error is already enough to judge it as failing; in a seven-page sample, however, seven such errors are far more than we would be willing to tolerate. Instead, we would prefer to fail the sample already at three or four major errors. This poses a problem for the linear scoring model which simply prorates the number of errors per page to a total number of pages in the sample.''
\end{quote}

Another industrial user confirmed these findings and chose to avoid the issue by only evaluating samples of exactly 1000 words. However, in real-world operations, such precisely consistent evaluation sample sizes are rare. Evaluation samples often vary significantly in length due to the nature of the content, deadlines, or operational constraints.

\section{Linear vs.\ Logarithmic Fits}
\label{appendix:linear-vs-logarithmic}

To complement the qualitative trend in Figures~\ref{fig:client1}--\ref{fig:client3}, we compare linear and logarithmic functional forms by fitting multiple candidate models to the same elicited tolerance points and evaluating goodness-of-fit. We report standard error-based measures (SSE, RMSE), explained variance (\(R^2\)), and information criteria (AIC, BIC), where lower values indicate better fit for SSE/RMSE/AIC/BIC and higher values indicate better fit for \(R^2\).

\paragraph{Fit metrics.}
For observed points \(\{(x_i,y_i)\}_{i=1}^{n}\) and fitted values \(\hat y_i\), the metrics are defined as:
\[
\mathrm{SSE}=\sum_{i=1}^{n}(y_i-\hat y_i)^2,\qquad
\mathrm{RMSE}=\sqrt{\frac{1}{n}\sum_{i=1}^{n}(y_i-\hat y_i)^2},
\]
\[
R^2=1-\frac{\sum_i (y_i-\hat y_i)^2}{\sum_i (y_i-\bar y)^2},\qquad
\mathrm{AIC}=n\ln(\mathrm{SSE}/n)+2k,\quad
\mathrm{BIC}=n\ln(\mathrm{SSE}/n)+k\ln n,
\]
where \(n\) is the number of points and \(k\) is the number of fitted parameters.

\paragraph{Models.}
We compare three candidate forms:
\begin{align*}
\text{(i) Logarithmic:} \quad & y = a\ln(1+bx), \\
\text{(ii) Linear (with intercept):} \quad & y = \alpha + \beta x, \\
\text{(iii) Linear (through origin):} \quad & y = cx.
\end{align*}
Model~(iii) is included because proportional tolerance scaling implicitly assumes a zero intercept (i.e., no tolerance at zero length), whereas Model~(ii) provides a standard unconstrained linear baseline.

Table~\ref{tab:fitquality3} reports fitted parameters and fit statistics for the Institution~1 minor-error elicitation dataset
(data points: \((2,2),(3,3),(4,4),(5,5),(7,6),(10,7),(20,8)\)).

\begin{table}[htbp]
\centering
\sisetup{
  group-digits=false,
  round-mode=places,
  round-precision=3,
  table-number-alignment = center
}
{\small
\setlength{\tabcolsep}{4pt}
\begin{tabular}{l
                S[table-format=1.3]
                S[table-format=1.3]
                S[table-format=1.3]
                S[table-format=1.3]
                S[table-format=2.3]
                S[table-format=1.3]
                S[table-format=1.3]
                S[table-format=-2.3]
                S[table-format=-2.3]}
\toprule
Model & {$a$} & {$b$} & {$\alpha$} & {$\beta$} & {SSE} & {RMSE} & {$R^2$} & {AIC} & {BIC} \\
\midrule
Logarithmic $a\ln(1+bx)$
 & 3.353 & 0.590 & {} & {} & 1.551 & 0.471 & 0.945 & -6.550 & -6.658 \\
Linear (with intercept) $\alpha+\beta x$
 & {} & {} & \text{---} & \text{---} & \text{---} & \text{---} & \text{---} & \text{---} & \text{---} \\
Linear (through origin) $cx$
 & {} & {} & {} & {} & 26.755 & 1.955 & 0.044 & 11.386 & 11.331 \\
\bottomrule
\end{tabular}
\par\medskip
\footnotesize
AIC $= n\ln(\mathrm{SSE}/n)+2k$, BIC $= n\ln(\mathrm{SSE}/n)+k\ln n$; here $n{=}7$.
Parameter counts: $k{=}2$ for logarithmic, $k{=}2$ for linear with intercept, and $k{=}1$ for linear through origin.
}
\caption{\textbf{Goodness-of-fit statistics} for Institution~1 minor-error elicitation (\(x\) in pages; \(y\) in minor-error equivalents).}
\label{tab:fitquality3}
\end{table}

\paragraph{Comparison.}
The logarithmic model substantially improves fit compared to a purely proportional linear rule. Reporting both constrained and unconstrained linear baselines strengthens the robustness of the comparison and avoids conflating functional form with parameter constraints.

\section{Calibration from Two Points (Numerical)}
\addcontentsline{toc}{section}{Appendix A. Calibration from Two Points (Numerical)}
\label{app:two-point-cal}

\indent \textbf{Checking the model feasibility.}

\smallskip
Given two tolerance points \((x_0,E_0)\), \((x_1,E_1)\) with \(x_0\neq x_1\), \(E_0,E_1>0\), define \(r=E_1/E_0\), \(\rho=x_1/x_0\). A unique calibration with \(a>0,b>0\) exists iff
\[
\min\{1,\rho\}<r<\max\{1,\rho\}.
\]

\smallskip

\textbf{One‑dimensional root‑finding.}
Let \(f(b)=\ln(1+b x_1)-r\ln(1+b x_0)\). Near the origin, \(f(b)\approx b(x_1-r x_0)\). As \(b\to\infty\), \(f(b)\sim (1-r)\ln b + \ln x_1 - r\ln x_0\), which flips sign relative to the origin when the feasibility condition holds. 
\smallskip

\textbf{Bisection (robust).}
\begin{enumerate}\itemsep0.25ex
\item Compute \(r=E_1/E_0\) and verify feasibility.
\item Set \(b_{\mathrm{lo}}\gets 0^+\) and evaluate \(f(b_{\mathrm{lo}})\).
\item Grow \(b_{\mathrm{hi}}\) geometrically until \(f(b_{\mathrm{lo}})\,f(b_{\mathrm{hi}})<0\).
\item Bisect until convergence: \(b\gets (b_{\mathrm{lo}}+b_{\mathrm{hi}})/2\) and update the bracket by the sign of \(f(b)\).
\item Output \(b^\star\) and \(a^\star=E_0/\ln(1+b^\star x_0)\).
\end{enumerate}

\noindent\emph{Notes.} (i) Changing the log base rescales $a$ only; we use the natural logarithm \(\ln\). (ii) For \(b\approx 0\), \(\ln(1+bx)\approx bx\), so the model reduces locally to a linear rule, clarifying why a single linear slope works only near its reference sample size. (iii) With \(n\ge 3\) points \(\{(x_i,E_i)\}\), fit \((a,b)\) by nonlinear least squares under \(a>0,b>0\).

\section{Least-Squares Calibration from Multiple Points}\label{app:ls-calibration}
\addcontentsline{toc}{section}{Appendix B. Least-Squares Calibration from Multiple Points}

\textbf{Problem.}
\smallskip

Given tolerance points \(\{(x_i,E_i)\}_{i=1}^{n}\) with \(x_i>0\) and \(E_i>0\), estimate \(a>0,b>0\) in
\[
E(x) \;=\; a\,\ln\!\bigl(1+b\,x\bigr)
\]
by (constrained) least squares:
\[
\min_{a>0,\;b>0}\; S(a,b)
\;\;:=\;\; \sum_{i=1}^{n}\Bigl[E_i - a\,L_i(b)\Bigr]^2,
\qquad L_i(b):=\ln\!\bigl(1+b\,x_i\bigr).
\]

\textbf{Profiling out \(a\).}
\smallskip

For any fixed \(b>0\), the minimizer in \(a\) is
\[
a(b) \;=\; \frac{\sum_i E_i\,L_i(b)}{\sum_i L_i^2(b)}.
\]
Substituting yields the one‑dimensional \emph{profiled} objective
\[
S(b)
\;=\; \sum_i E_i^2 \;-\; \frac{\Bigl(\sum_i E_i\,L_i(b)\Bigr)^{\!2}}{\sum_i L_i^2(b)}
\;=\; S_0 - \frac{S_1(b)^2}{S_2(b)},
\]
where \(S_0=\sum_i E_i^2\), \(S_1(b)=\sum_i E_i\,L_i(b)\), and \(S_2(b)=\sum_i L_i^2(b)\).\\

\textbf{Derivative for 1‑D solvers (optional).}
\smallskip

Let \(L_i'(b)=\dfrac{x_i}{1+b\,x_i}\).
Then
\[
S_1'(b)=\sum_i E_i\,L_i'(b), \qquad
S_2'(b)=2\sum_i L_i(b)\,L_i'(b),
\]
and
\[
S'(b) \;=\; -\,\frac{2\,S_1(b)\,S_1'(b)\,S_2(b) \;-\; S_1(b)^2\,S_2'(b)}{S_2(b)^2}.
\]
While not strictly required, \(S'(b)\) enables Newton/Brent methods. In practice, \(S(b)\) is
unimodal over \(b>0\) for the elicited data we observed.\\

\textbf{Numerically stable 1‑D procedure.}
\smallskip
\begin{enumerate}
  \item If a feasible two‑point calibration is available (~\nameref{app:two-point-cal}),
        use its \(b^\star\) as the starting value; otherwise set \(b_0 = 1/(10\,\max_i x_i)\).
  \item Define \(q(b)=S(b)\). Bracket a minimum by expanding geometrically from \(b_0\)
        (e.g., multiply by \(2\)) until \(q\) increases on both sides.
  \item Minimize \(q(b)\) on the bracket with a derivative‑free method (Brent or golden‑section).
  \item Set \(\hat b=\arg\min q(b)\) and \(\hat a=a(\hat b)\).
\end{enumerate}

\textbf{Weighted least squares (optional).}
\smallskip

If different points have different reliabilities, use positive weights \(w_i\) to minimize
\[
S_w(a,b)=\sum_i w_i\bigl[E_i - a\,L_i(b)\bigr]^2
\]
with
\[
a_w(b)=\frac{\sum_i w_i E_i\,L_i(b)}{\sum_i w_i L_i^2(b)},\qquad
S_w(b)=\sum_i w_i E_i^2 - \frac{\bigl(\sum_i w_i E_i L_i(b)\bigr)^2}{\sum_i w_i L_i^2(b)}.
\]
Then proceed as above with \(S_w(b)\).\\

\textbf{Approximate standard errors.}
\smallskip

Let residuals \(r_i=E_i-\hat a\,L_i(\hat b)\) and \(\hat\sigma^2=S(\hat a,\hat b)/(n-2)\).
Define the Jacobian \(J\in\mathbb{R}^{n\times 2}\) at \((\hat a,\hat b)\):
\[
\frac{\partial r_i}{\partial a}=-L_i(\hat b),\qquad
\frac{\partial r_i}{\partial b}=-\,\hat a\,L_i'(\hat b) \;=\; -\,\hat a\,\frac{x_i}{1+\hat b\,x_i}.
\]
Then an approximate covariance for \((\hat a,\hat b)\) is
\[
\widehat{\mathrm{Cov}}(\hat a,\hat b)\;=\;\hat\sigma^2\,(J^\top J)^{-1}.
\]
For any \(x>0\), the delta method gives the variance of predicted tolerance
\[
\widehat{\mathrm{Var}}\!\left[E(x)\right] \;\approx\;
\nabla_\theta E(x)^\top\,\widehat{\mathrm{Cov}}(\hat a,\hat b)\,\nabla_\theta E(x),
\quad
\nabla_\theta E(x)=\begin{bmatrix}
\ln(1+\hat b\,x) \\[3pt]
\hat a\,\dfrac{x}{1+\hat b\,x}
\end{bmatrix}.
\]
A nonparametric alternative is a bootstrap over the elicited points.\\

\textbf{Implementation notes.}
\smallskip

(i) Use \(\ln\) for the natural log and enforce \(b>0\). (ii) Rescale \(x\) if needed to keep
\(1+\hat b\,x\) well‑conditioned numerically. (iii) When points are very near two‑point feasible,
the LS minimum occurs close to the two‑point solution.

\textbf{Weights and ribbons.} If point reliabilities differ, use weights \(w_i>0\). Approximate covariance of \((\hat a,\hat b)\) and delta‑method variance of predictions \(E(x)\) are given above; use these to compute pointwise standard errors and plot confidence ribbons around the fitted curve.

\textbf{Sensitivity to anchors (illustrative).}
With the second point fixed at \((x_1,E_1)=(250,2)\), varying the primary anchor by \(\pm 1\) at \(x_0=1000\) yields:

\begin{table}[h]
\centering
\begin{tabular}{@{}lccc@{}}
\toprule
Anchor \(E_0\) & \(E(1000)\) & \(E(2000)\) & \(E(3000)\) \\
\midrule
4 & 4.00 & 5.16 & 5.86 \\
\textbf{5 (baseline)} & \textbf{5.00} & \textbf{7.05} & \textbf{8.36} \\
6 & 6.00 & 9.30 & 11.59 \\
\bottomrule
\end{tabular}
\end{table}

Near the anchor, predictions are stable by construction; farther away, two‑point calibrations diverge—hence the recommendation to use least squares with multiple points when available (this Appendix~B).


\section{Excel Goal Seek method for Two-Point Calibration}
\addcontentsline{toc}{section}{Appendix C. Excel Goal Seek method}

\begin{enumerate}
    \item Put $x_0$ in \textbf{A1}, $E_0$ in \textbf{B1}, $x_1$ in \textbf{C1}, and $E_1$ in \textbf{D1}.
    \item Enter an initial guess for $b>0$ in \textbf{E1}.
    \item Compute $\varphi(b)$ in cell \textbf{F1} using the formula:
    \[
    \texttt{=LN(1 + \$E\$1 * \$C\$1) - (\$D\$1 / \$B\$1) * LN(1 + \$E\$1 * \$A\$1)}
    \]
    \item Go to \textit{Data} $\rightarrow$ \textit{What-If Analysis} $\rightarrow$ \textit{Goal Seek}: set cell \textbf{F1} to 0 by changing cell \textbf{E1}.
    \item Once Goal Seek converges, compute $a$ in cell \textbf{G1} using the formula:
    \[
    \texttt{=\$B\$1 / LN(1 + \$E\$1 * \$A\$1)}
    \]
\end{enumerate}


\section{Minimal Python code for Two-Point Calibration}
\addcontentsline{toc}{section}{Appendix D. Minimal Python code for Two-Point Calibration}

Below is a minimal two-point calibration function that computes the non-linear tolerance
parameters \(a\) (overall scale) and \(b\) (curvature) as described in the paper.

\begin{verbatim}
import math

def two_point_calibrate(x0, E0, x1, E1):
    r = E1 / E0
    def f(b):  # score-equation for b
        return math.log(1.0 + b*x1) - r*math.log(1.0 + b*x0)

    # bracket a root
    blo = 1e-12
    f_lo = f(blo)
    bhi = 1e-6
    for _ in range(80):
        f_hi = f(bhi)
        if f_lo * f_hi < 0:
            break
        bhi *= 2.0
    else:
        raise ValueError("Could not bracket a root for b; check inputs.")

    # bisection
    for _ in range(120):
        bmid = 0.5 * (blo + bhi)
        f_mid = f(bmid)
        if abs(f_mid) < 1e-12:
            blo = bhi = bmid
            break
        if f_lo * f_mid < 0:
            bhi = bmid
        else:
            blo = bmid
            f_lo = f_mid

    b = 0.5 * (blo + bhi)
    a = E0 / math.log(1.0 + b*x0)
    return a, b
\end{verbatim}

{Command-line usage:}
Save as \texttt{calibrate.py} and append the wrapper below to call it from a terminal:

\begin{verbatim}
if __name__ == "__main__":
    import argparse, math
    p = argparse.ArgumentParser(
        description="Two-point calibration for E(x)=a*ln(1+b*x)")
    p.add_argument("--x0", type=float, required=True, help="reference size x0")
    p.add_argument("--E0", type=float, required=True, help="tolerance at x0")
    p.add_argument("--x1", type=float, required=True, help="second size x1")
    p.add_argument("--E1", type=float, required=True, help="tolerance at x1")
    p.add_argument("--x",  type=float, help="optional size to evaluate E(x)")
    args = p.parse_args()

    a, b = two_point_calibrate(args.x0, args.E0, args.x1, args.E1)
    print(f"a={a:.6f}, b={b:.8g}")
    if args.x is not None:
        E = a * math.log(1.0 + b*args.x)
        print(f"E({args.x:g})={E:.6f}")
\end{verbatim}

\noindent\textbf{Example 1 (reproduce §5.3).}
\begin{verbatim}
python3 calibrate.py --x0 1000 --E0 5 --x1 250 --E1 2
\end{verbatim}
Expected output (approx.): \texttt{a=3.688000, b=0.00288}

\noindent\textbf{Example 2 (compute \(E(3{,}000)\), §7.3).}
\begin{verbatim}
python3 calibrate.py --x0 1000 --E0 5 --x1 250 --E1 2 --x 3000
\end{verbatim}
Expected output (approx.): \texttt{E(3000)=8.357}


\section{Least-Squares Fit (multiple points)}
\addcontentsline{toc}{section}{Appendix E. Least-Squares Fit (multiple points)}

This routine fits the logarithmic tolerance \(E(x)=a\,\ln(1+bx)\) to \emph{multiple}
tolerance points \(\{(x_i,E_i)\}_{i=1}^n\) by profiling out \(a\) and minimizing the
one-dimensional objective in \(b\) (see App.~B). Given any \(b>0\),
\[
a(b)=\frac{\sum_i E_i \ln(1+b x_i)}{\sum_i \ln^2(1+b x_i)}\!,
\quad
S(b)=\sum_i\!\left(E_i-a(b)\ln(1+b x_i)\right)^2,
\]
and we choose \(\hat b=\arg\min_{b>0} S(b)\), then \(\hat a=a(\hat b)\).

Python code for Least Squares Fit:

\begin{verbatim}
import math

def calibrate_log_lsq(points, weights=None, b0=None, grow=2.0, tol=1e-8):
    """
    Least-squares fit of E(x) = a * ln(1 + b x) to multiple points.
    points  : list of (x, E) pairs with x>0, E>0
    weights : optional list of positive weights w_i (same length as points)
    b0      : optional initial guess for b; default 1 / (10 * max x)
    returns : (a_hat, b_hat, sse)
    """

    xs, Es = zip(*points)
    n = len(xs)
    if weights is None:
        ws = [1.0] * n
    else:
        if len(weights) != n:
            raise ValueError("weights must match points length")
        ws = list(weights)

    xmax = max(xs)
    if b0 is None or b0 <= 0:
        b0 = 1.0 / (10.0 * xmax)

    def L(b):
        # vector of ln(1 + b x_i); return None if invalid
        try:
            return [math.log(1.0 + b * x) for x in xs]
        except ValueError:
            return None

    def a_of_b(b):
        Li = L(b)
        if Li is None or min(1.0 + b * x for x in xs) <= 0:
            return None
        num = sum(w * E * l for w, E, l in zip(ws, Es, Li))
        den = sum(w * l * l for w, l in zip(ws, Li))
        if den <= 0:
            return None
        return num / den

    def sse_profile(b):
        a = a_of_b(b)
        if a is None:
            return float("inf")
        Li = L(b)
        r = [math.sqrt(w) * (E - a * l) for w, E, l in zip(ws, Es, Li)]
        return sum(ri * ri for ri in r)

    # ---- bracket a minimum around b0 by geometric expansion
    a_l = b0 / grow
    a_c = b0
    a_r = b0 * grow
    f_l = sse_profile(a_l)
    f_c = sse_profile(a_c)
    f_r = sse_profile(a_r)

    # ensure f_c is the smallest
    it = 0
    while not (f_c <= f_l and f_c <= f_r) and it < 80:
        it += 1
        if f_l < f_r:
            a_r, f_r = a_c, f_c
            a_c, f_c = a_l, f_l
            a_l /= grow
            f_l = sse_profile(a_l)
        else:
            a_l, f_l = a_c, f_c
            a_c, f_c = a_r, f_r
            a_r *= grow
            f_r = sse_profile(a_r)

    if it == 80:
        raise ValueError("Could not bracket a minimum for b; check data.")

    # ---- golden-section search on [a_l, a_r]
    phi = (math.sqrt(5) - 1) / 2
    left, right = a_l, a_r
    x1 = right - phi * (right - left)
    x2 = left + phi * (right - left)
    f1 = sse_profile(x1)
    f2 = sse_profile(x2)

    while (right - left) > tol * (abs(left) + abs(right) + 1.0):
        if f1 > f2:
            left = x1
            x1, f1 = x2, f2
            x2 = left + phi * (right - left)
            f2 = sse_profile(x2)
        else:
            right = x2
            x2, f2 = x1, f1
            x1 = right - phi * (right - left)
            f1 = sse_profile(x1)

    b_hat = 0.5 * (left + right)
    a_hat = a_of_b(b_hat)
    sse = sse_profile(b_hat)
    return a_hat, b_hat, sse
\end{verbatim}

To test this code from the command line, add the wrapper below (or place it in a separate \texttt{lsq\_fit.py}) to fit from a list of
\((x,E)\) pairs and optionally evaluate \(E(x)\) at a target size.

\begin{verbatim}
if __name__ == "__main__":
    import argparse, math, sys
    p = argparse.ArgumentParser(
        description="Least-squares fit for E(x)=a*ln(1+b*x)")
    p.add_argument("--xy", action="append", metavar="X,E",
                   help="data point as 'x,E' (repeat for multiple points)")
    p.add_argument("--x", type=float, help="optional size to evaluate E(x)")
    args = p.parse_args()

    if not args.xy:
        sys.exit("Provide at least one --xy 'x,E' pair.")
    pts = []
    for pair in args.xy:
        x_str, E_str = pair.split(",")
        pts.append((float(x_str), float(E_str)))

    a, b, sse = calibrate_log_lsq(pts)
    print(f"a={a:.6f}, b={b:.8g}, SSE={sse:.6f}")
    if args.x is not None:
        print(f"E({args.x:g})={a*math.log(1.0+b*args.x):.6f}")
\end{verbatim}

\noindent\textbf{Example 1 (user~1, pages vs.\ minor errors; Fig.~3/Table~1).}
\begin{verbatim}
python3 lsq_fit.py \
  --xy 2,2 --xy 3,3 --xy 4,4 --xy 5,5 --xy 7,6 --xy 10,7 --xy 20,8
\end{verbatim}

Expected output (approx.): \texttt{a=3.353, b=0.59046, SSE}$\approx$\texttt{1.551}.\\

\noindent\textbf{Example 2 (evaluate the fitted curve at 12 pages).}
\begin{verbatim}
python3 lsq_fit.py \
  --xy 2,2 --xy 3,3 --xy 4,4 --xy 5,5 --xy 7,6 --xy 10,7 --xy 20,8 \
  --x 12
\end{verbatim}
Expected output (approx.): \texttt{E(12)}$\approx$7.22.


\section{Wolfram Notebook for Least Squares Fit}
\addcontentsline{toc}{section}{Appendix E. Wolfram Notebook for Least Squares Fit}

This appendix provides a Wolfram Notebook implementation of the least-squares
calibration method described in Section~5.3 and detailed in Appendix~A.
The code can be pasted directly into a Wolfram Notebook and run cell by cell.
It reproduces the profiled least-squares fitting procedure for the model
\[
E(x) \;=\; a\,\ln\!\bigl(1+b\,x\bigr), \qquad a>0, \; b>0,
\]
using bracketing and golden-section search over $b$, with $a$ profiled in closed form.

\subsection{Instructions}
\begin{enumerate}
  \item Open Wolfram Mathematica and create a new Notebook.
  \item Copy the code blocks below into separate cells.
  \item Evaluate the cells in order (Shift+Enter).
  \item The final cell produces a plot and fitted formula.
\end{enumerate}

\subsection*{Cell 1: Definitions}

\begin{verbatim}
ClearAll[CalibrateLogLSQ];
Options[CalibrateLogLSQ] = {"Weights"->None,"InitialB"->Automatic,
   "BracketStep"->2.,"Tolerance"->1.*^-8};

CalibrateLogLSQ[data_List, opts:OptionsPattern[]] := Module[
  {x,E,n,wOpt,w,b0,step,tol,L,Lp,aOfB,sseProfile,bracketMin,
   goldenMin,br,bHat,aHat,yHat,resid,sse,sigma2,J,JTJ,cov},
  x=N[data[[All,1]]]; E=N[data[[All,2]]]; n=Length[x];
  If[AnyTrue[x,#<=0&]||AnyTrue[E,#<=0&],Return[$Failed,Module]];
  wOpt=OptionValue["Weights"];
  w=If[wOpt===None,ConstantArray[1.,n],N[wOpt]];
  If[Length[w]=!=n||AnyTrue[w,#<=0&],Return[$Failed,Module]];
  b0=Replace[OptionValue["InitialB"],Automatic->(1./(10. Max[x]))];
  step=OptionValue["BracketStep"]; tol=OptionValue["Tolerance"];

  L[b_?NumericQ]:=Log[1.+b*x];
  Lp[b_?NumericQ]:=x/(1.+b*x);

  aOfB[b_?NumericQ]:=Module[{Li=L[b],num,den},
    If[Min[1.+b*x]<=0,Return[Indeterminate]];
    num=Total[w*E*Li]; den=Total[w*Li*Li];
    If[den<=0||!NumericQ[den],Indeterminate,num/den]];

  sseProfile[b_?NumericQ]:=Module[{a=aOfB[b],Li,r},
    If[!NumericQ[a],Return[Infinity]];
    Li=L[b]; r=Sqrt[w]*(E-a*Li); r.r];

  bracketMin[f_,bStart_,s_:2.,maxIter_:60]:=Module[{a,b,c,fa,fb,fc,it=0},
    a=bStart/s; b=bStart; c=bStart*s; fa=f[a]; fb=f[b]; fc=f[c];
    While[!(fb<fa&&fb<fc)&&it<maxIter,
      it++;
      If[fa<fb,
        c=b;fc=fb; b=a;fb=fa; a=a/s; If[a<=0,a=b/(2 s)]; fa=f[a],
        a=b;fa=fb; b=c;fb=fc; c=c*s; fc=f[c]]];
    {a,b,c}];

  goldenMin[f_,a0_,c0_,t_:1.*^-8,maxIter_:200]:=Module[
    {phi=(Sqrt[5]-1)/2.,a=a0,c=c0,x1,x2,f1,f2,it=0},
    x1=c-phi(c-a); x2=a+phi(c-a); f1=f[x1]; f2=f[x2];
    While[(c-a)>t(Abs[a]+Abs[c]+1.)&&it<maxIter,
      it++;
      If[f1>f2, a=x1; x1=x2; f1=f2; x2=a+phi(c-a); f2=f[x2],
                 c=x2; x2=x1; f2=f1; x1=c-phi(c-a); f1=f[x1]]];
    .5(a+c)];

  br=bracketMin[sseProfile,b0,step];
  bHat=goldenMin[sseProfile,br[[1]],br[[3]],tol];
  aHat=aOfB[bHat]; If[!NumericQ[aHat],Return[$Failed,Module]];

  yHat=aHat*L[bHat]; resid=E-yHat;
  sse=(Sqrt[w]*resid).(Sqrt[w]*resid);
  sigma2=sse/Max[n-2,1];
  J=Transpose[{-L[bHat],-aHat*Lp[bHat]}];
  JTJ=If[wOpt===None,Transpose[J].J,Transpose[Sqrt[w]*J].(Sqrt[w]*J)];
  cov=sigma2*PseudoInverse[JTJ];

  <|"a"->aHat,"b"->bHat,"SSE"->sse,"Sigma2"->sigma2,
    "Residuals"->resid,"Covariance"->cov,
    "Predict"->(aHat*Log[1.+bHat*#]&),"Bracket"->br[[{1,3}]]|>
];
\end{verbatim}

\subsection*{Cell 2: Fit and Quick Results}

\begin{verbatim}
data = {{2,2},{3,3},{4,4},{5,5},{7,6},{10,7},{20,8}};
fit  = CalibrateLogLSQ[data];
{fit["a"], fit["b"], fit["SSE"]} // N
\end{verbatim}

\subsection*{Cell 3: Plot}

\begin{verbatim}
 With[{a = fit["a"], b = fit["b"]},
 Show[
   {
     ListPlot[data, PlotStyle -> {Blue, PointSize[.02]}],
     ListLinePlot[data, InterpolationOrder -> 1,
                  PlotStyle -> {Blue, Dotted, Thick}],
     Plot[a*Log[1 + b*x], {x, 0.5, 25}, PlotStyle -> {Red, Thick}]
   },
   GridLines -> Automatic, Frame -> True,
   FrameLabel -> {"Pages (x)", "Tolerance  E(x)"},
   PlotLabel -> Style["Least-squares log fit", 14, Bold],
  Epilog -> {
  Inset[
    Style[
      Row[{"E(x) \[TildeTilde] ", NumberForm[a, {5,5}], " ln(1 + ",
           NumberForm[b, {5,5}], " x)"}],
      12, Bold, Red],
    Scaled[{0.6, 0.65}]
  ]
 },
   ImageSize -> Large
 ]
]
\end{verbatim}

\subsection{Illustration and Resulting Formula}

\begin{figure}[h]
  \centering
  \includegraphics[width=1\linewidth]{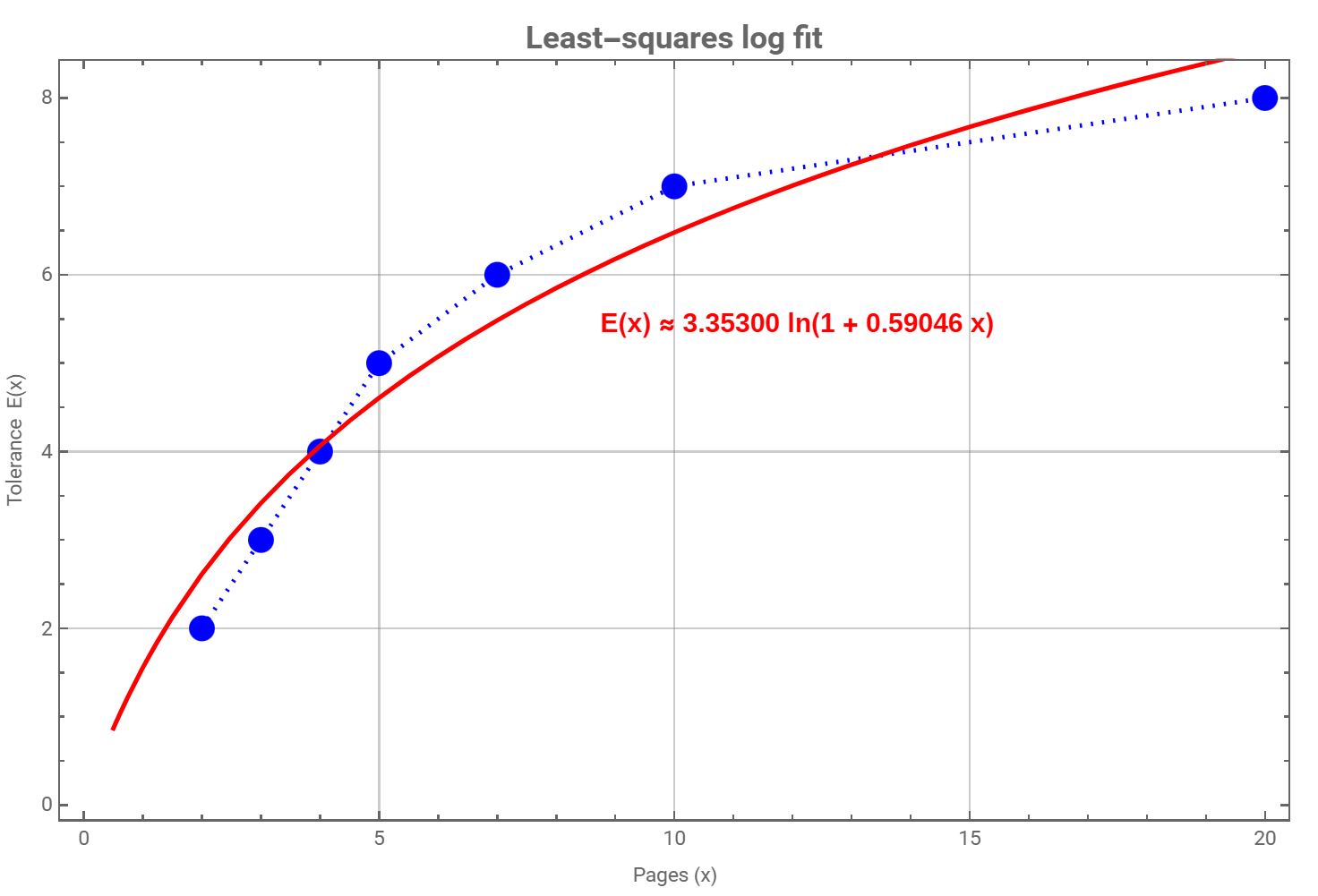}
  \caption{Least-squares logarithmic fit in Wolfram Notebook.}
  \label{fig:wolfram-lsq}
\end{figure}

Figure~\ref{fig:wolfram-lsq} shows the illustration produced by Cell~3.
The fitted formula is returned by \verb|{fit["a"], fit["b"]}|
in Cell~2, which evaluates numerically to
\[
E(x) \;\approx\; \hat a \,\ln\!\bigl(1 + \hat b\,x\bigr),
\]
with coefficients $\hat a, \hat b$ depending on the dataset.

For the sample dataset serving as example, the approximated function is:
\[
E(x) \;\approx\; 3.35301 \,\ln\!\bigl(1 + 0.59046\,x\bigr),
\]
with residual sum of squares (SSE) $\approx 1.55087$.

For different dataset change \textit{data} array in Cell 2.

\end{appendices}

\bibliography{sn-bibliography}
\end{document}